\documentclass[10pt,journal,compsoc]{IEEEtran}
\ifCLASSOPTIONcompsoc
  \usepackage[nocompress]{cite}
\else
  \usepackage{cite}
\fi
\ifCLASSINFOpdf
\else
\fi
\usepackage{url}
\usepackage{amsmath}
\usepackage{amssymb}
\usepackage{multirow}
\usepackage{booktabs}
\usepackage{listings}
\usepackage{algorithm} 
\usepackage{graphicx}
\usepackage{color}
\usepackage{subcaption}
\usepackage{makecell}

\usepackage{enumitem}
\usepackage{ragged2e} % for justifying
\usepackage[pagebackref=false,breaklinks=true,colorlinks,bookmarks=false]{hyperref}

\newcommand{\oursattn}{SSA\xspace}
\newcommand{\oursabbrv}{Spikformer\xspace}
\newcommand{\newoursabbr}{Spikformer~V2\xspace}

\newcommand{\oursfull}{Spiking Transformer\xspace}

\usepackage{amsfonts}       % blackboard math symbols
\usepackage{nicefrac}       % compact symbols for 1/2, etc.
\usepackage{microtype}      % microtypography
\usepackage{xcolor}         % colors
\usepackage{xspace}
\usepackage{multirow}
\usepackage{adjustbox}
\usepackage{pifont}
\usepackage{bbding}
\usepackage{fontawesome} 
\newcommand{\cmark}{\ding{51}}%
\newcommand{\xmark}{\ding{55}}%
\newcommand{\tabincell}[2]{\begin{tabular}{@{}#1@{}}#2\end{tabular}}
\usepackage{tikz}
\usepackage{soul}

\usepackage[noend]{algpseudocode}
\usepackage{algorithmicx,algorithm}
\usepackage{tabularx}
\usepackage{tikz}
\usepackage{multirow}
\usepackage[mathscr]{euscript}
\usepackage{enumitem}
\definecolor{natural}{rgb}{0.7137,0.3333,0.3333}
\definecolor{specialized}{rgb}{0.4118,0.6431,0.4314}
\definecolor{structured}{rgb}{0.3254,0.4431,0.6666}
\definecolor{all}{rgb}{0.7529,0.4902,0.6471}
\definecolor{alexey}{rgb}{0.8, 0.0, 0.8}

\definecolor{matthias}{rgb}{0.0, 0.8, 0.8}

\definecolor{sylvain}{rgb}{0.8, 0.8, 0.0}

\newcommand{\op}[1]{\operatorname{#1}}

\begin{document}
\title{Spikformer V2: Join the High Accuracy Club on ImageNet with an SNN Ticket}

\author{Zhaokun Zhou,
        Yijie Lu,
        Kaiwei Che,
        Wei Fang,
        Keyu Tian,
        Qihao Peng,\\
        Yuesheng Zhu,
        Shuicheng Yan,~\IEEEmembership{Fellow,~IEEE,}
        Yonghong Tian,~\IEEEmembership{Fellow,~IEEE,}
        and Li Yuan
\IEEEcompsocitemizethanks{
% \IEEEcompsocthanksitem M. Shell was with the Department
% of Electrical and Computer Engineering, Georgia Institute of Technology, Atlanta,
% GA, 30332.\protect\\
% E-mail: see http://www.michaelshell.org/contact.html
% \IEEEcompsocthanksitem J. Doe and J. Doe are with Anonymous University.}% <-this % stops an unwanted space
% \IEEEcompsocthanksitem $^\dagger$ means Co contributions.
\IEEEcompsocthanksitem Zhaokun Zhou, Yijie Lu, Kaiwei Che, Yuesheng Zhu, Yonghong Tian and Li Yuan are with Peking University, School of Electronic and Computer Engineering, Shenzhen Graduate School, China, and also with PengCheng Laboratory. E-mail: \{zhouzhaokun, chekaiwei\}@stu.pku.edu.cn, and \{zhuys, yhtian, yuanli-ece\}@pku.edu.cn.
\IEEEcompsocthanksitem Wei Fang is with Peking University, School of Computer Science, China, and also with PengCheng Laboratory. E-mail: {fwei}@pku.edu.cn.
\IEEEcompsocthanksitem Keyu Tian is with Peking University, Center for Data Science, China. E-mail: {keyutiankeyutian}@stu.pku.edu.cn.
\IEEEcompsocthanksitem Qihao Peng is with the 5G and 6G Innovation Centre, Institute for Communication Systems (ICS), University of Surrey, Guildford, GU2 7XH, U.K. E-mail: q.peng@surrey.ac.uk. Qihao Peng is with the 5G and 6G Innovation Centre, Institute for Communication Systems (ICS), University of Surrey, Guildford, GU2 7XH, U.K. E-mail: {q.peng}@surrey.ac.uk.
\IEEEcompsocthanksitem Shuicheng Yan is with Beijing Academy of Artificial Intelligence. E-mail: {shuicheng.yan}@gmail.com.
% \IEEEcompsocthanksitem This work is extended from our conference version in ICLR\cite{zhou2023spikformer}.
}
\thanks{Manuscript received August 31, 2023.}}

\definecolor{psnncolor}{HTML}{FFA09D}
\definecolor{spikformercolor}{HTML}{FAAC8E}
\definecolor{newspikformercolor}{HTML}{776BAA}
\definecolor{vitcolor}{HTML}{AEC6FA}
\newcommand{\psnncolor}[1]{\textcolor{psnncolor}{#1}}
\newcommand{\spikformercolor}[1]{\textcolor{spikformercolor}{#1}}
\newcommand{\newspikformercolor}[1]{\textcolor{newspikformercolor}{#1}}
\newcommand{\vitcolor}[1]{\textcolor{vitcolor}{#1}}
\newcommand{\spb}{\spikformercolor{$\bullet$\,}}
\newcommand{\nspb}{\newspikformercolor{$\bullet$\,}}
\newcommand{\psnnb}{\psnncolor{$\bullet$\,}}
\newcommand{\vb}{\vitcolor{$\bullet$\,}}
\newcommand{\gr}{\rowcolor[gray]{.95}}

\IEEEtitleabstractindextext{

\justifying 

\begin{abstract}
Spiking Neural Networks (SNNs), known for their biologically plausible architecture, face the challenge of limited performance. 
The self-attention mechanism, which is the cornerstone of the high-performance Transformer and also a biologically inspired structure, is absent in existing SNNs. 
To this end, we explore the potential of leveraging both self-attention capability and biological properties of SNNs, and propose a novel Spiking Self-Attention (SSA) and Spiking Transformer (Spikformer). 
The SSA mechanism eliminates the need for softmax and captures the sparse visual feature employing spike-based Query, Key, and Value. This sparse computation without multiplication makes SSA efficient and energy-saving. 
Further, we develop a Spiking Convolutional Stem (SCS) with supplementary convolutional layers to enhance the architecture of Spikformer.
The \oursabbrv enhanced with the SCS is referred to as \newoursabbr. 
To train larger and deeper \newoursabbr, we introduce a pioneering exploration of Self-Supervised Learning (SSL) within the SNN.
Specifically, we pre-train \newoursabbr with masking and reconstruction style inspired by the mainstream self-supervised Transformer, and then finetune the \newoursabbr on the image classification on ImageNet. 
Extensive experiments show that \newoursabbr outperforms other previous surrogate training and ANN2SNN methods. 
An 8-layer \newoursabbr achieves an accuracy of 80.38\% using 4 time steps, and after SSL, a 172M 16-layer \newoursabbr reaches an accuracy of 81.10\% with just 1 time step. To the best of our knowledge, this is the first time that the SNN achieves 80+\% accuracy on ImageNet.
The code will be available at \href{https://github.com/ZK-Zhou/spikformer}{\newoursabbr}.

\end{abstract}

}

\maketitle

\IEEEdisplaynontitleabstractindextext

% \fi
\IEEEpeerreviewmaketitle

\IEEEraisesectionheading{\section{Introduction}\label{sec:introduction}}

\IEEEPARstart{H}uman brains with incredible efficiency and the ability to perform complex pattern recognition tasks have been objects of imitation and sources of innovation for deep Artificial Neural Networks (ANNs).
In recent years, deep learning has developed rapidly with ANNs and achieved remarkable success in many fields, including natural language processing \cite{devlin2018bert,brown2020language,touvron2023llama}, computer vision \cite{dosovitskiy2020image, he2022masked,kirillov2023segment}, and robotics \cite{brohan2022rt}. 
Exploring how to leverage the intrinsic efficient computation paradigm of biological neural systems to implement neural networks with low power consumption is of great value.
Spiking Neural Networks (SNNs) are a promising way to achieve energy-efficient intelligence, as they mimic biological neuronal behavior by using spiking signals for inter-neuron communication \cite{maass1997networks}. 
SNNs can run smoothly on neuromorphic chips \cite{merolla2014million,davies2018loihi,pei2019towards} by only performing spike-based accumulation operations and skipping zero values of input or activation (i.e., event-driven) \cite{roy2019towards,NEURIPS2021_61f2585b}, which consume much less power than ANNs.

\begin{figure}[]
    \centering
    \includegraphics[width=0.9\linewidth]{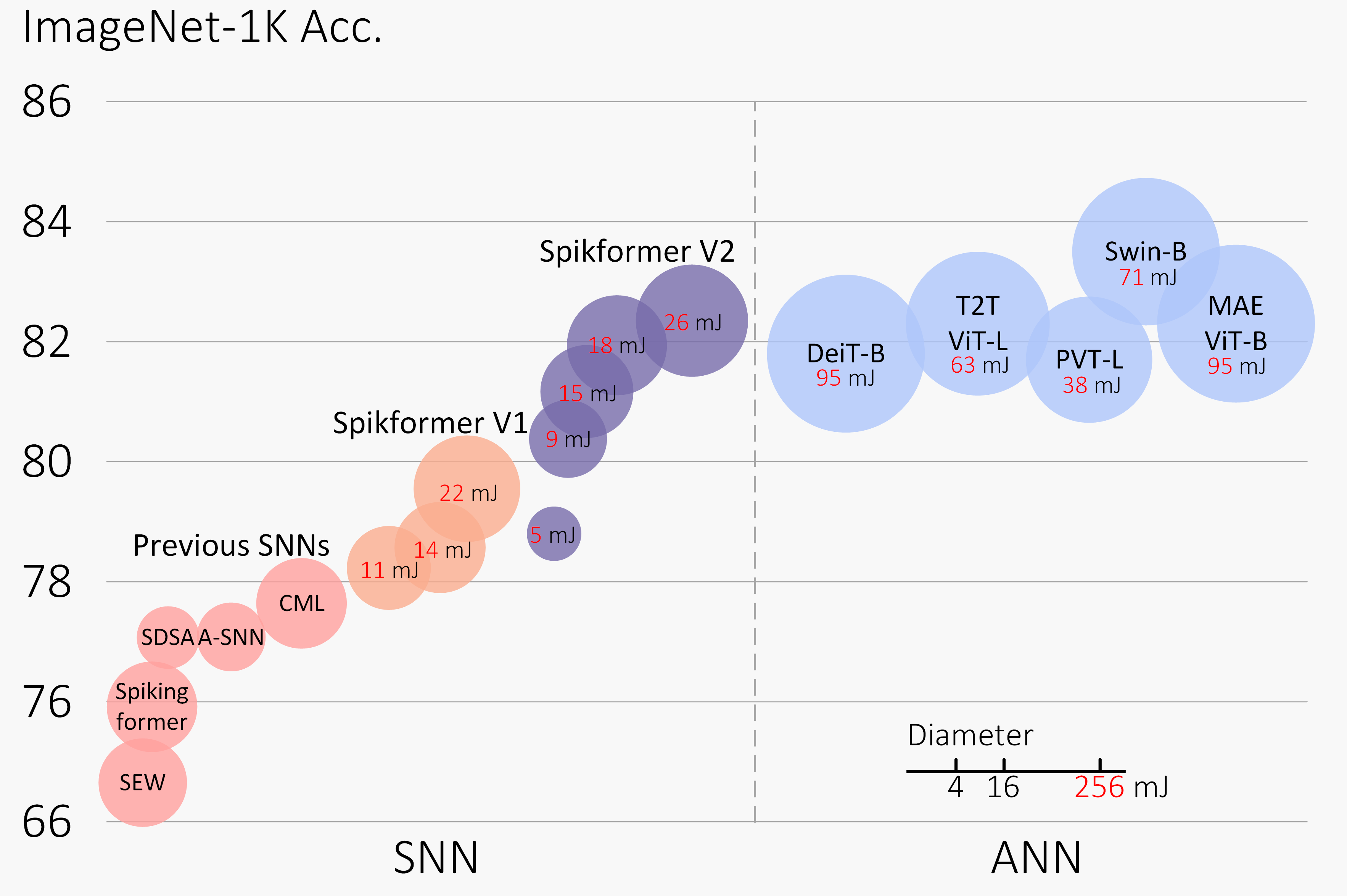}
    \caption{ImageNet-1K classification results for \psnnb previous SNNs, \spb \oursabbrv V1, \nspb \newoursabbr and \vb Vision Transformers. The diameter of each bubble is logarithmically proportional to the theoretical energy consumption.
    We demonstrate that the \newoursabbr can achieve equivalent classification accuracy levels to those of ANN-Transformers, while maintaining lower theoretical energy consumption.}
    \label{fig:circle}
\end{figure}

Training large-scale SNNs to achieve competitive performance in real-world pattern recognition tasks still presents a significant challenge.
Some works attempt to build larger-scale networks from various perspectives, including the design of spiking residual networks \cite{fang2021deep,hu2021advancing}, attention mechanisms \cite{yao2021temporal}, spiking neuron designs \cite{fang2021incorporating, cheng2023meta}.
However, in ANNs, as shown in Fig. \ref{fig:circle}, the classification accuracy on ImageNet \cite{deng2009imagenet} generally exceeds 80\%~\cite{dosovitskiy2021an,touvron2021deit,yuan2021tokens,yuan2021volo,wang2021pyramid,liu2021swin,chen2021pre}, with the highest reaching 91.1\% \cite{chen2023symbolic}. 
The existing directly trained pure SNNs usually achieve below 80\%\cite{hu2021advancing,zheng2021going,fang2021deep}, with the highest being 78.46\% \cite{zhou2023enhancing}. 
To overcome the bottleneck, we need to construct more powerful and efficient SNN.
\begin{figure*}[tb]
    \centering
    \includegraphics[width=1.0\linewidth]{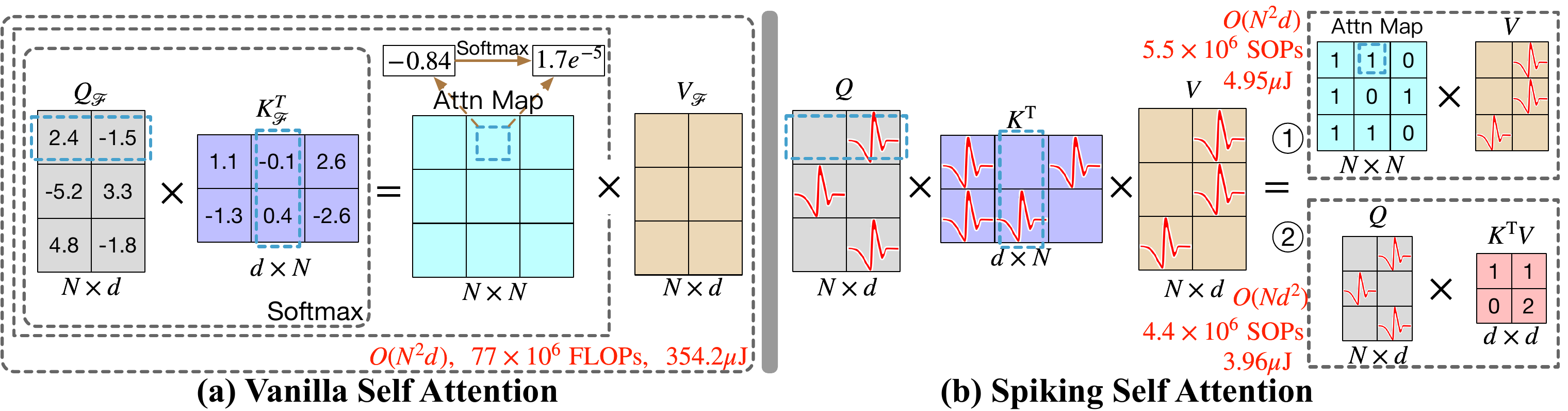}
    \caption{
    Comparison between Vanilla Self-Attention (VSA) and our Spiking Self-Attention (\oursattn).
    A red spike indicates a value of 1 at a specific location. The blue dashed boxes demonstrate examples of matrix dot product operations.
    For simplicity, we select one of the heads of \oursattn, where $N$ represents the number of input patches and $d$ denotes the feature dimension of one head.
    FLOPs stands for floating-point operations, and SOPs represents the theoretical synaptic operations.
    The theoretical energy consumption for performing one calculation between Query, Key, and Value in one time step is derived from the 8-encoder-blocks 512-embedding-dimension \oursabbrv on the test set of ImageNet, using the method in \cite{kundu2021hire, hu2018residual,zhou2023spikformer}.
    (a) In VSA, $Q_{\mathcal{F}},K_{\mathcal{F}},V_{\mathcal{F}}$ are in float-point forms. 
    After the dot-product of $Q_{\mathcal{F}}$ and $K_{\mathcal{F}}$, the softmax function regularizes the attention map values to be positive.
    (b) In \oursattn, all values in the attention map are non-negative and the computation is sparse using spike-form $Q, K, V$ ($5.5\times 10^6$ VS. $77 \times 10^6$ in VSA).
    As a result, \oursattn consumes less energy compared to VSA ({$354.2\mu \rm{J}$}). 
    The \oursattn is decomposable (the calculation order of $Q,K$ and $V$ is changeable).
    }
    \label{fig:ssa}
\end{figure*}
Self-attention can select interest information by modeling both long and short dependencies of input features, enabling Transformer to flourish in various domains of deep learning, including Natural Language Processing (NLP)\cite{vaswani2017attention,devlin2018bert}, computer vision \cite{dosovitskiy2020image,liu2021swin,yuan2021volo}, and large language/vision models\cite{touvron2023llama,brown2020language,kirillov2023segment}.
However, the exploration of self-attention mechanisms in SNNs is still absent. Given the biological inspiration behind these two mechanisms and the excellent performance of self-attention, applying self-attention to SNNs for more advanced deep learning is an intriguing avenue to explore.

However, integrating the self-attention mechanism into SNNs is not straightforward.
Vanilla Self-Attention (VSA) \cite{vaswani2017attention} consists of three components: Query, Key, and Value. As illustrated in Fig. \ref{fig:ssa}~(a), 
The standard procedure of VSA computes a floating-point matrix through Query and Key dot product, followed by softmax normalization (involving exponentiation and division) to create the attention map for Value scaling. These steps clash with the computational characteristics of SNNs, i.e., multiplication avoidance.
Furthermore, the high computational cost of VSA (dense floating-point calculations) makes it impractical to apply to SNNs directly.
Hence, to enable Transformer on SNNs, we require a new self-attention variant that is effective, computation-efficient, and multiplication-free.
We thus propose Spiking Self-Attention (SSA), as shown in Fig. \ref{fig:ssa}~(b).
SSA is the first to apply the self-attention mechanism to SNNs, capturing interdependence using spike sequences.
In SSA, the Query, Key, and Value are in spike form which only consists of 0 and 1. The challenges to the adoption of self-attention in SNNs are mainly due to softmax.
As depicted in Fig. \ref{fig:ssa}, the attention map computed from spike-form Query and Key has inherent non-negativeness, filtering out irrelevant features.
Therefore, we do not require the softmax to ensure the non-negative attention matrix, which is its primary role in VSA \cite{qin2022cosformer}.
Table \ref{tab:ablation1} demonstrates that our SSA is comparable to VSA in the effect of processing spike sequences.
Based on these insights, we eliminate softmax normalization for the attention map in SSA.
Some prior Transformer variants also eliminate softmax or substitute it with a linear function.
For instance, in Performer\cite{choromanski2020rethinking}, the positive random feature is used to approximate softmax; CosFormer\cite{qin2022cosformer} replaces softmax with ReLU and cosine function.

The sparse spike-form Query, Key, and Value in \oursattn are computed through logical AND operations and addition without involving multiplications, resulting in efficiency due to sparsity and simplified operations. 
Based on the \oursattn, we propose the \oursfull (\oursabbrv). Fig. \ref{fig:spikformer} shows the overview of \oursabbrv. 
It attains high performance on both static and neuromorphic datasets. To the best of our knowledge, this is the first work to investigate the self-attention and directly-training Transformer in SNNs.

Despite the improvement over previous SNNs, \oursabbrv still lags significantly behind ANN performance on ImageNet.
Two common approaches to enhancing \oursabbrv performance are optimizing network architecture and improving learning methods.
Regarding architecture improvement, we eliminate the max-pooling in Spike Patch Splitting module that might cause information loss and blur features.
Inspired by the substantial benefits of early-stage injection of convolutional induced biases in ANN Vision Transformer (ViT)\cite{xiao2021early}, we explore the significance of shallow convolutions in SNN-Transformer and the impact of the number of convolution layers on model performance, as depicted in Fig. \ref{fig:scs_conv}.
The redesigned patch splitting module is referred to as the Spiking Convolutional Stem (SCS) and the \oursabbrv based on SCS is named \newoursabbr.

Despite the performance improvement achieved by \newoursabbr, we observe that training becomes unstable and performance degrades when \newoursabbr exceeds more than 12 layers. 
To obtain larger and deeper \newoursabbr, we investigate the masked image modeling \cite{he2022masked,bao2022beit}, a Self-Supervised Learning method that demonstrates efficient and stable training on large ANN-Transformers.
However, Self-Supervised Learning, an important learning mode in the brain \cite{zhuang2021unsupervised,zhuang2022well}, has not been extensively studied in SNNs.
We pre-train \newoursabbr with Mask Autoencoder style \cite{he2022masked}.
We use the standard ANN-Transformer as a decoder to better reconstruct the masked images and obtain a powerful \newoursabbr. 
Note that using the ANN-Transformer decoder does not affect \newoursabbr as a pure SNN.
It will be discarded after pre-training, and only the \newoursabbr will be used for subsequent fine-tuning.
With SSL, we can train larger and deeper \newoursabbr, which further unleashes the potential of SSA and Spikformer V2 and elevate their performance upper bound.

As shwon in Fig. \ref{fig:circle}, after supervised training, \newoursabbr-8-512 attains an accuracy of 80.38\% across 4 simulation time steps.
Based on self-supervised pre-training, \newoursabbr-16-768 achieves 81.10\% accuracy on ImageNet with only 1 time step.
To the best of our knowledge, this is the first time that SNNs surpass 80\% on ImageNet.
This further emphasizes the potential of large-scale SNNs to achieve high performance, in addition to their biocompatibility and computational efficiency.

The rest of this paper is structured as follows. Sec.~\ref{sec:2} introduces the background and related work. Next, we review \oursabbrv and its components, particularly Spiking Self-Attention, in Sec.~\ref{sec:3}. We provide a detailed description of \newoursabbr in Sec.~\ref{sec:4}. Subsequently, we present the experimental results in Sec.~\ref{sec:5}. Finally, we conclude and discuss future work in Sec.~\ref{sec:6}.

\section{Background and Related Work} \label{sec:2}

\textbf{Vision Transformers.}
Transformers, initially designed for NLP tasks \cite{vaswani2017attention,devlin2018bert}, have demonstrated impressive performance in computer vision, including image classification \cite{dosovitskiy2020image,yuan2021tokens}, object detection \cite{carion2020end,zhu2020deformable,liu2021swin}, semantic segmentation \cite{wang2021pyramid,yuan2021volo}, and low-level image processing \cite{chen2021pre}.
In the context of image classification, a conventional ViT \cite{dosovitskiy2020image} architecture comprises three main components: a patch splitting module, transformer encoder layers, and a linear classification head. The transformer encoder layer encompasses a self-attention block and a multi-perception layer block. Self-attention is a fundamental component that plays a crucial role in the success of ViT. 
It assigns weights to the feature Values of image patches based on the dot-product of Query and Key, followed by a softmax function. 
It enables the model to capture global dependencies and represent important patterns of interest in the input data \cite{katharopoulos2020transformers,qin2022cosformer}.
Several studies have explored enhancements to the structure of ViT. One notable improvement is the utilization of convolution layers for patch splitting, which has demonstrated the ability to expedite convergence and mitigate the data-intensive nature of ViT \cite{xiao2021early,hassani2021escaping}.
Subsequent studies, such as PVT~\cite{DBLP:journals/corr/abs-2102-12122}, Swin~\cite{DBLP:journals/corr/abs-2103-14030}, and MViTv2~\cite{Li2021MViTv2IM}, have integrated the pyramid structure with transformers and eliminated the class token originally present in the architecture.
Various approaches are proposed to address the computational complexity of self-attention and enhance its capacity to model visual dependencies \cite{song2021ufo,yang2021focal,rao2021dynamicvit,choromanski2020rethinking}. In this study, we specifically investigate the efficacy of self-attention in SNNs and aim to develop a powerful Spiking Transformer for image classification.

\textbf{Self-Supervised Learning.}
In ANNs, Self-Supervised Learning (SSL) has emerged as a promising paradigm for both computer vision and natural language processing \cite{he2022masked,bao2022beit,devlin2018bert}. 
Contrastive learning and mask auto-encoding are two mainstream SSL.
Contrastive learning \cite{chen2020simple,he2020momentum,caron2021emerging,Chen_2021_ICCV} aims to learn invariances by comparing augmented views of unlabeled images.
Mask auto-encoding enables the learning of robust representations by reconstructing masked inputs using simple data augmentation techniques \cite{devlin2018bert, bao2022beit, he2022masked}.
BEiT \cite{bao2022beit} pioneered the integration of mask auto-encoding into the computer vision community, while MAE \cite{he2022masked} further refined this approach by introducing an asymmetric encoder and decoder architecture. In MAE, the encoder selectively bypasses masked tokens to alleviate computational burden, while the decoder processes all tokens.
We will explore mask auto-encoding in SNNs for the first time.

\textbf{Spiking Neural Networks.} In contrast to conventional deep learning models that utilize continuous decimal values to represent information, SNNs employ discrete spike sequences for computation and information transmission.
Spiking neurons, such as the Leaky Integrate-and-Fire (LIF) neuron \cite{wu2018spatio}, PLIF \cite{fang2021incorporating}, and others, are capable of converting continuous input signals into spike sequences.
There are two approaches to obtaining deep SNN models: ANN-to-SNN conversion and direct training.
In the process of ANN-to-SNN conversion\cite{cao2015spiking,hunsberger2015spiking,rueckauer2017conversion,wu2021progressive,meng2022training,wang2022signed}, the ReLU activation layers in a high-performance pre-trained ANN are replaced with spiking neurons to obtain an equivalent SNN model.
The converted SNN requires a larger number of time steps to accurately approximate the ReLU activation function, resulting in increased latency \cite{han2020rmp}.
In the field of direct training \cite{fang2021deep,zhu2023spikegpt}, SNNs are unfolded over a series of simulation time steps and trained using a variant of the back-propagation through time algorithm \cite{lee2016training,shrestha2018slayer}.
Due to the non-differentiable nature of the event-triggered mechanism in spiking neurons, surrogate gradients are employed for back-propagation\cite{lee2020enabling,neftci2019surrogate}. In \cite{xiao2021training}, implicit differentiation on the equilibrium state is adopted to train SNNs.
Various models originally developed for ANNs have been successfully transferred to SNNs. However, the investigation of self-attention in SNNs remains an open research area. Convolution-based temporal attention was proposed as a means to reduce unnecessary time steps in SNNs \cite{yao2021temporal}.
Both \cite{zhang2022spiking,zhang2022spike} employ ANN-Transformer architectures for spike data processing, despite their titles containing  "Spiking Transformer". \cite{mueller2021spiking} presents an ANN-SNN conversion Transformer, but it still utilizes vanilla self-attention. In this study, we aim to investigate the feasibility of integrating self-attention and Transformer into SNNs.

\section{\oursabbrv} \label{sec:3}

\begin{figure*}[t]
    \centering
    \includegraphics[width=0.8\linewidth]{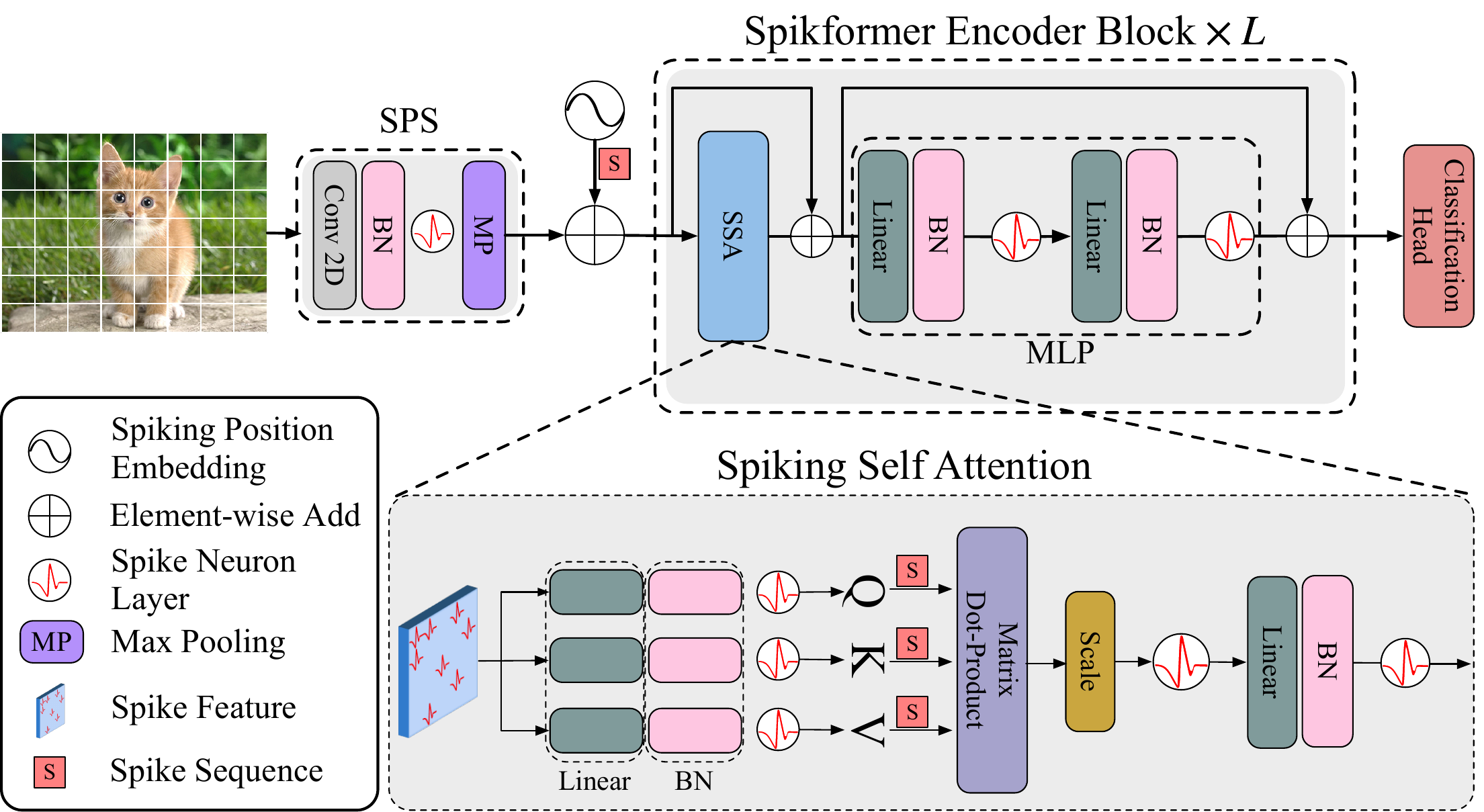}
    \caption{
        The architecture of \oursfull (\oursabbrv) includes a spiking patch splitting module (SPS), a \oursabbrv encoder, and a Linear classification head. We observe that layer normalization (LN) is not suitable for SNNs, hence we empirically utilize batch normalization (BN) instead.
        }
    \label{fig:spikformer}
\end{figure*}

We introduce \oursfull~(\oursabbrv), a pure SNN that integrates the self-attention and Transformer for improved learning capability. Next, we present the overview and components of \oursabbrv in detail.

\subsection{Overall Architecture} \label{sec:spikformer_overview}
Fig. \ref{fig:spikformer} shows the structure and components of \oursabbrv. The Spiking Patch Splitting (SPS) module converts a 2D image sequence $I\in \mathbb{R}^{T \times C\times H\times W}$\footnote{The data shape in the neuromorphic dataset is $I\in \mathbb{R}^{T \times C\times H\times W}$, where $T$, $C$, $H$, and $ W$ represent time step, channel, height and width, respectively. A 2D image $I_s \in \mathbb{R}^{C\times H\times W}$ in static datasets has to be duplicated $T$ times to form a sequence of images.} into a $D$ dimensional spike-form feature vector and segments it into a sequence of $N$ flattened spike-form patches $x$ using linear projections. SNNs do not support float-point-form position embedding. We use a conditional position embedding generator\cite{chu2021twins} to produce spike-form relative position embedding (RPE) and add the RPE to patches sequence $x$ to obtain $X_0$. 
The conditional position embedding generator is composed of a 2D convolution layer (Conv2d) with a kernel size of 3, followed by batch normalization (BN), and a spiking neuron layer ($\mathcal{SN}$).

We feed the $X_0$ to the $L$-block \oursabbrv encoder. A \oursabbrv encoder block contains a Spiking Self Attention (SSA) block and a Multi-Layer Perceptron (MLP) block, similar to the standard ViT encoder block. Residual connections are used in both the SSA and MLP blocks. As the main component in \oursabbrv encoder block, \oursattn provides an efficient way to model the local-global information of images using spike-form Query ($Q$), Key ($K$), and Value ($V$) without softmax, which will be discussed in detail in Sec. \ref{sec:ssa}. A Global Average-Pooling (GAP) is applied to the processed feature from \oursabbrv encoder and outputs the $D$ dimensional feature which will be sent to the fully-connected-layer Classification Head (CH) to output the prediction $Y$.
\oursabbrv can be expressed as follows:
\begin{align}
    x={\rm{SPS}}\left(I\right), {{I}} \in \mathbb{R}^{T \times C\times H\times W}, x\in \mathbb{R}^{T\times N\times D}.
\end{align}
\begin{align}
{\rm{RPE}}={\mathcal{SN}}({\rm{BN}}(({\rm{Conv2d}}(x)))), {\rm{RPE}}\in \mathbb{R}^{T \times N\times D}.
\end{align}
\begin{align}
     X_0 = x + {\rm{RPE}}, X_0\in \mathbb{R}^{T \times N\times D}.
\end{align}
\begin{align}
X^{\prime}_l = {\rm{SSA}}(X_{l-1}) + X_{l-1}, X^{\prime}_l\in \mathbb{R}^{T \times N\times D},l=1...L .
\end{align}
\begin{align}
X_l = {\rm{MLP}}(X^{\prime}_l) + X^{\prime}_l, X_l\in \mathbb{R}^{T \times N\times D}, l=1...L .
\end{align}
\begin{align}
     X_0 = x + {\rm{RPE}}, X_0\in \mathbb{R}^{T \times N\times D}.
\end{align}

\subsection{Spiking Patch Splitting (SPS)}\label{sec:patch_transformer}
As shown in Fig. \ref{fig:spikformer}, the Spiking Patch Splitting (SPS) module is designed to transform an image into a spike-form feature of dimension $D$ and segment it into fixed-size patches. Multiple blocks can be included in the SPS module. Following the convolutional stem in Vision Transformer~\cite{xiao2021early,hassani2021escaping}, we employ a convolution layer in each block of SPS to inject inductive bias into \oursabbrv. Specifically, for an image sequence $I\in \mathbb{R}^{T \times C\times H\times W}$:
\begin{align}
x={\mathscr{MP}}\left({\mathcal{SN}}({\rm{BN}}(({\rm{Conv2d}}(I))))\right),
\end{align}
where the 2D convolution layer (stride-1, $3\times 3$ kernel size) and max-pooling are represented by Conv2d and $\mathscr{MP}$, respectively.
The SPS module can consist of more than 1 block. 
When there are multiple SPS blocks, the output channels of the convolution layers are exponentially increased until they equal the patch embedding dimension. 
For example, for an output embedding dimension $D$ and a four-block SPS module, the output channels of the four convolution layers are $D/8,D/4,D/2,D$. 
The 2D-max-pooling layer is applied to reduce the feature size with a fixed size after each SPS block. 
After the SPS processing, $I$ is converted into a sequence of image patches  $x\in \mathbb{R}^{ T \times N\times D}$.

\subsection{Spiking Self Attention Mechanism (SSA)}\label{sec:ssa}
The \oursabbrv encoder is the core component of the whole architecture, which comprises the Spiking Self Attention (SSA) mechanism and MLP block. We focus on SSA, beginning with a review of vanilla self-attention (VSA). For an input feature sequence $X \in \mathbb{R}^{T \times N\times D}$, the VSA in ViT has three float-point key components, namely Query ($Q_{\mathcal{F}}$), Key ($K_{\mathcal{F}}$), and Value ($V_{\mathcal{F}}$) which are computed by learnable linear matrices $W_Q, W_K, W_V \in\mathbb{R}^{D\times D }$ and $X$:
\begin{align}
Q_{\mathcal{F}} = XW_Q,~ K_{\mathcal{F}}=XW_K,~ V_{\mathcal{F}} = XW_V,
\end{align}
where ${\mathcal{F}}$ indicates the float-point form. The output of vanilla self-attention can be calculated as:
\begin{align}\label{eq:vsa}
{\rm{VSA}}(Q_{\mathcal{F}},K_{\mathcal{F}},V_{\mathcal{F}})={\rm{Softmax}}\left(\frac{Q_{\mathcal{F}}K_{\mathcal{F}}^{\rm{T}}}{\sqrt{d}}\right)V_{\mathcal{F}},
\end{align}
where $d={D}/{H}$ is the feature dimension of one head and $H$ is the head number. The transformation of the Value in float-point form ($V_{\mathcal{F}}$) to spike form ($V$) enables the direct use of VSA in SNNs, which can be written as:
\begin{align}\label{eq:vsainsnn}
{\rm{VSA}}(Q_{\mathcal{F}},K_{\mathcal{F}},V)={\rm{Softmax}}\left(\frac{Q_{\mathcal{F}}K_{\mathcal{F}}^{\rm{T}}}{\sqrt{d}}\right)V.
\end{align}
However, VSA is not compatible with SNNs for two reasons. 1) The float-point matrix multiplication of $Q_{\mathcal{F}}, K_{\mathcal{F}}$ and softmax function that involves exponential calculation and division operation, violate the computation rules of SNNs.
2) The quadratic space and time complexity of the sequence length of VSA do not satisfy the efficient computational demands of SNNs.

We introduce Spiking Self-Attention (\oursattn), which is more adapted to SNNs than VSA, as shown in Fig. \ref{fig:ssa}(b) and the bottom of Fig. \ref{fig:spikformer}. The Query, Key, and Value, which are denoted by $Q, K$, and $V$ respectively, are first obtained by applying learnable matrices. Then they are transformed into spiking sequences by different spike neuron layers:
\begin{align}\label{eq:spikeqkv}
&Q = {{\mathcal{SN}}_Q}(\op{BN}(XW_Q)),\nonumber \\
&K={{\mathcal{SN}_K}}(\op{BN}(XW_K)), \\
&V = {{\mathcal{SN}_V}}(\op{BN}(XW_V)), \nonumber
\end{align}
where $Q,K,V \in \mathbb{R}^{T \times N\times D}$. We argue that the computation process of the attention matrix should employ pure spike-form Query and Key (only containing 0 and 1). Motivated by vanilla self-attention \cite{vaswani2017attention}, we introduce a scaling factor $s$ to regulate the large value of the matrix multiplication result. $s$ does not alter the property of \oursattn, and as shown in Fig. \ref{fig:spikformer}, the spike-friendly \oursattn is formulated as:
\begin{align}
{\rm{\oursattn}}^{'}(Q,K,V)={\mathcal{SN}}\left({Q}~{K^{\rm{T}}}~V * s\right).
\label{eq:ssa}
\end{align}
The ${\mathcal{SN}}(\op{BN}(\op{Linear}(\cdot)))$ is used to transform and fuse the channel information\cite{vaswani2017attention}:
\begin{align}
{\rm{\oursattn}}(Q,K,V)={\mathcal{SN}}(\op{BN}(\op{Linear}({\rm{\oursattn}}^{'}(Q,K,V)))).
\label{eq:ssa2}
\end{align}

We can generalize the single-head \oursattn presented here to the multi-head \oursattn.
\oursattn operates on each time step. 
As shown in Eq.~(\ref{eq:ssa}), SSA eliminates the softmax normalization of the attention matrix in Eq.~(\ref{eq:vsa}) and directly multiplies $Q,K$ and $V$. Fig. \ref{fig:ssa}(b) illustrates a simple calculation example. Our \oursattn does not require softmax, which actually impedes the implementation of self-attention in SNNs. 
According to Eq. (\ref{eq:spikeqkv}), the spiking neuron layer  ${{\mathcal{SN}}_Q}$ and ${{\mathcal{SN}_K}}$ produce the spike sequences $Q$ and $K$, which are inherently non-negative (0 or 1). This leads to a non-negative attention map. SSA filters out the irrelevant information and only retains the relevant features. Hence, the softmax is not required to guarantee the non-negativeness of the attention map. 

In contrast to the float-point-form $X_{\mathcal{F}}$ and $V_{\mathcal{F}}$ in ANNs, the spike form of the input $X$ and the Value $V$ of self-attention in SNNs has limited information content. The spike-form coarse features $X,V$ in SNNs do not require the fine-grained modeling of VSA, which uses float-point-form $Q_{\mathcal{F}}$,$K_{\mathcal{F}}$ and softmax operations that are redundant and ineffective for extracting more information from $X,V$ than \oursattn. That is, \oursattn is a more appropriate and efficient self-attention mechanism for SNNs than VSA.
\begin{table}[]
\centering
\caption{Analyzing the rationality of \oursattn.
We substitute \oursattn with alternative attention variants while keeping the remaining network structure in \oursabbrv unchanged. 
We show the accuracy (Acc) on CIFAR10-DVS \cite{li2017cifar10}, CIFAR10/100 \cite{krizhevsky2009learning}. OPs (M) is the number of operations (For $\rm{A_I},\rm{A_{LeakyReLU}},\rm{A_{ReLU}}$ and $\rm{A_{softmax}}$, OPs is FLOPs, and SOPs is ignored; For $\rm{A_{\oursattn}}$, it is SOPs.) and $\rm P$ ($\mu \rm{J}$) is the theoretical energy consumption for a single calculation involving $Q,K,V$.}
\scalebox{1.0}{
    \setlength{\tabcolsep}{0.9mm}{
        \begin{tabular}{lccc}
                \toprule
  &  {CIFAR10-DVS} & {CIFAR10} &{CIFAR100} \\
\midrule
  &&  {Acc/OPs (M)/P ($\mu {\rm{J}}$)} & \\
%   {Acc/SOPs/P($\mu \rm{J}$)} &{Acc/OPs/P($\mu \rm{J}$)}
\midrule
$\rm{A_I}$  &79.40/16.8/{77} &  93.96/6.3/{29} & 76.94/6.3/{29} \\
\midrule
$\rm{A_{LeakyReLU}}$   &79.80/16.8/{77}	&93.85/6.3/{29} &76.73/6.3/{29}     \\
\midrule
$\rm{A_{ReLU}}$   &79.40/{16.8/77} &  94.34/{6.3/29} & 77.00/{6.3/29}       \\
\midrule
$\rm{A_{softmax}}$ & 80.00/19.1/{88} &  94.97/6.6/{30} & \textbf{77.92}/6.6/{30} \\
\midrule
$\rm{A_{\oursattn}}$  & \textbf{80.90/0.66/{0.594}} &  \textbf{95.19}/\textbf{1.1}/{\textbf{0.990}} & 
{77.86}/\textbf{1.3}/{\textbf{1.170}}\\
                \bottomrule
\end{tabular}}}
\label{tab:ablation1}
\end{table}
To verify the above observations, we compare our \oursattn with four other attention map computation methods in the experiments, as illustrated in Tab. \ref{tab:ablation1}. $\rm{A_I}$ represents the attention map derived from directly multiplying the float-points  $Q_{\mathcal{F}}$ and $K_{\mathcal{F}}$, which keeps both positive and negative correlation. $\rm{A_{ReLU}}$ employs the multiplication of ${\rm{ReLU}}(Q_{\mathcal{F}})$ and ${\rm{ReLU}}(K_{\mathcal{F}})$ to generate the attention map.  $\rm{A_{ReLU}}$ maintains the positive values of $Q_{\mathcal{F}}$, $K_{\mathcal{F}}$ and zeros out the negative values, while  $\rm{A_{LeakyReLU}}$ preserves the negative values. $\rm{A_{softmax}}$ indicates the attention map is computed following VSA. The above four methods adopt the same \oursabbrv framework and apply weights to the spike-form $V$.

As shown in Table \ref{tab:ablation1}, our $\rm{A_{SSA}}$ outperforms $\rm{A_I}$ and $\rm{A_{LeakyReLU}}$, demonstrating the advantage of computing attention map using ${\mathcal{SN}}$. %naturally retaining positive values.
We speculate that $\rm{A_{SSA}}$ surpasses $\rm{A_{ReLU}}$  because $\rm{A_{SSA}}$ exhibits stronger non-linearity in self-attention. Through comparative analysis with $\rm{A_{softmax}}$, our proposed model, $\rm{A_{\oursattn}}$, exhibits competitive performance and even outperforms $\rm{A_{softmax}}$ in terms of CIFAR10DVS and CIFAR10 datasets. The superiority of \oursattn over VSA in handling spike sequences ($X$ and $V$) with restricted information can be ascribed to its inherent adaptability. Additionally, the computational complexity and theoretical energy consumption associated with the calculation of $Q, K, V$ by $\rm{A_{SSA}}$ are considerably reduced compared to alternative methods.

\oursattn is specifically tailored for the purpose of modeling spike sequences. By representing $Q, K$, and $V$ in spike form, the matrix dot-product calculation is simplified to logical AND and summation operations. For instance, considering a row of Query $q$ and a column of Key $k$, the calculation $\sum_{i=1}^{d}{q_i}{k_i}=\sum_{q_i=1}k_i$ illustrates this transformation. Furthermore, as depicted in Table \ref{tab:ablation1}, our proposed model, \oursattn, exhibits reduced computational burden and energy consumption attributed to the sparsity of spike-form $Q, K$, and $V$ (as illustrated in Fig. \ref{fig:ssa}) and the simplified calculations. Moreover, the order of computation between $Q, K$, and $V$ is interchangeable: either performing $QK^{\rm{T}}$ first followed by $V$, or conducting $K^{\rm{T}}V$ first followed by $Q$. In scenarios where the sequence length $N$ exceeds the dimensionality $d$ of a single head, the latter computation order incurs lower computational complexity ($O(Nd^2)$) compared to the former ($O(N^2d)$). \oursattn preserves both the biological plausibility and computational efficiency properties consistently throughout the entire calculation process.

% We provide the pseudo-code of LSDA in the appendix (\ref{apd:pesudo}). Based on the vanilla multi-head self-attention, LSDA can be implemented with only ten lines of code. Further, only \textit{reshape} and \textit{permute} operations are used, so no extra computational cost is introduced.

\section{\newoursabbr} \label{sec:4}
In this section, we propose a Spiking Convolutional Stem (SCS) to replace the original SPS, aiming to alleviate the issues of blurry spike block features and the inability to perform mask pre-training. Additionally, Self-Supervised Learning for \oursabbrv is explored to further enhance the performance of SNNs on ImageNet. With the introduction of the enhanced SCS and Self-Supervised Learning, an updated version of the model that is more focused on larger-scale models is obtained, referred to as \newoursabbr.
\begin{figure}[]
\centering
\includegraphics[width=0.8\linewidth]{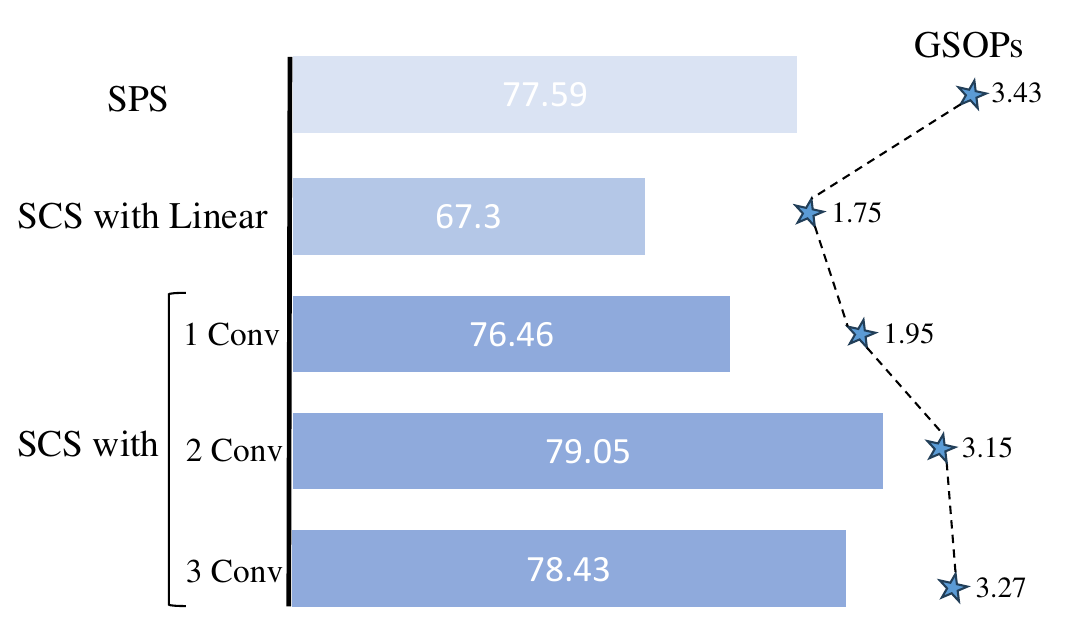}
\caption{Redesign of patch-splitting module. SPS represents the ImageNet accuracy achieved using \oursabbrv with SPS. 
When we remove max-polling, the accuracy of SCS with Linear layers sees a significant drop (the second row), whereas SCS with Convolutional layers performed better and reached its peak with the two-layer convolution configuration. 
All models have approximately the same number of parameters.
GSOPs represents the theoretical synaptic operations.}
\label{fig:scs_conv}
\end{figure}
\begin{figure}[]
\centering
\includegraphics[width=0.7\linewidth]{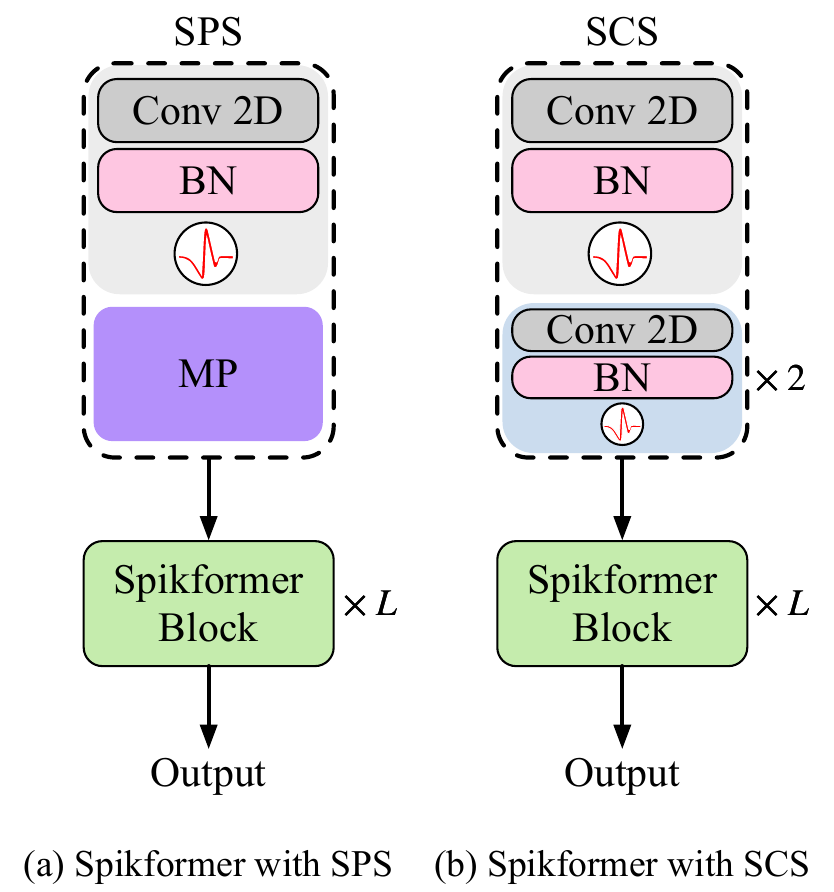}
\caption{Comparison between the Spiking Patch Splitting (SPS) and the Spiking Convolutional Stem (SCS). (a) In each block of the SPS module, the initial operation involves applying a 2D convolution with a kernel size of 3 and a stride of 1. This is followed by a max-pooling operation for downsampling by a factor of 2, which may lead to a potential loss of feature information. (b)
In each block of the SCS module, the initial step involves downsampling using a 2D convolution with a kernel size of 2 and a stride of 2. Additionally, we augment each SCS block with a convolutional block structure similar to the MLP Block.}
\label{fig:scs}
\end{figure}

\begin{figure*}[]
    \centering
    \includegraphics[width=0.9\linewidth]{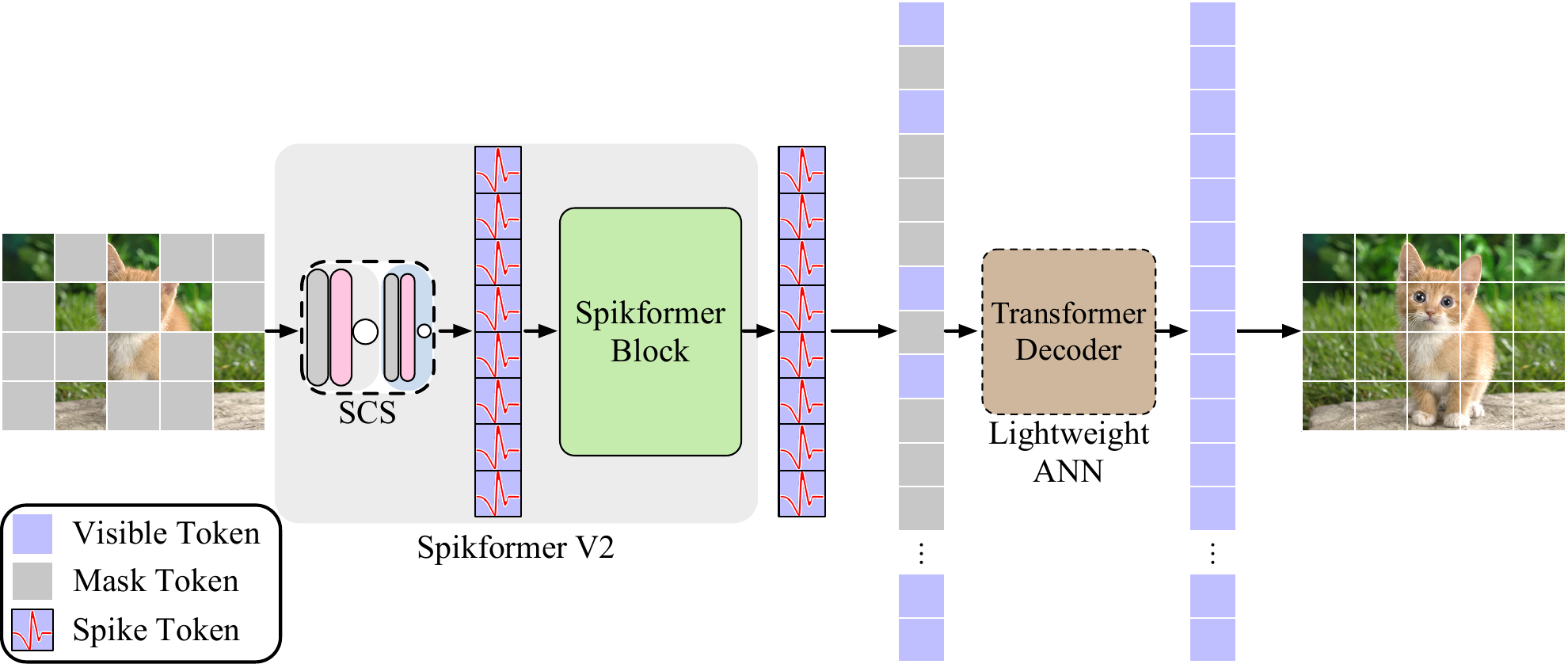}
    \caption{Self-supervised pre-training on \newoursabbr using the Mask Auto-Encoder framework.
    The SCS spike tokens are randomly masked using a high mask ratio.
    \newoursabbr performs feature modeling on visible image patches and outputs spike tokens, while Transformer reconstructs the images using floating-point mask tokens and visible tokens.
    To prevent information leakage from masked patches, we adopt sparse convolution and sparse batch normalization in SCS, following the approach used in MC-MAE \cite{gao2022mcmae} and SparK \cite{tian2023designing}.
    The Transformer is a lightweight network ($3.41M$) that is discarded after pre-training.
    }
    \label{fig:spikmae}
\end{figure*}

\subsection{Spiking Convolutional Stem (SCS)}\label{section:SCS}
Prior research \cite{zhou2023spikformer} identified the importance of the SPS module in \oursabbrv, which integrates convolution and progressive down-sampling. Despite performing well on small datasets, the model performance on ImageNet is not competitive. 
Additionally, the \oursabbrv using the SPS module cannot gain performance benefits from self-supervised pre-training.
As indicated in Table \ref{tab:spikmae}, the result of the Spikformer employing SPS after SSL and fine-tuning is lower than that of supervised training (75.23\% VS. 78.23\%).
We attribute this phenomenon to two main reasons: 1)  The SPS comprises four 'Conv-BN-LIF-MP' blocks, with each block containing a max-pooling layer. Continuous max-pooling tends to result in a significant loss of information in spike-based features, and causes blurring of the features after SPS block mapping. Another study \cite{zhou2023enhancing} has indicated that SPS can lead to erroneous gradient back-propagation. 
2) In the earlier patch splitting module, stacking convolutions can greatly improve optimization stability and enhance peak performance \cite{xiao2021early}. Despite the presence of 1 convolutional layer in each SPS block, it is still insufficient to fully optimize the performance of the model. 
As illustrated in Fig. \ref{fig:scs_conv}, upon removing max-pooling, the addition of linear layers within SCS leads to a sharp decline in performance, while adding convolutional layers (with a kernel size of 3, stride of 1, and padding of 1) results in a performance boost.
To balance performance and efficiency, two additional convolutional layers are introduced within each SCS block.

Inspired by the aforementioned insights, we propose a redesigned patch splitting module for Spikformer, referred to as the Spiking Convolutional Stem (SCS) Module, which is shown in Fig. \ref{fig:scs}. 
This module differs from the SPS module in two main aspects. Firstly, instead of utilizing max-pooling for downsampling, SCS employs standard 2D convolution \cite{dosovitskiy2020image}, thereby avoiding information loss in feature representation. 
Secondly, we introduce an increased number of convolution layers, where two consecutive layers form a group.
Existing studies \cite{zhang2023hivit} have suggested that the MLP blocks play a crucial role in learning shallow-level features.
Therefore, we structure these two convolutional layers to resemble MLP blocks, enabling SCS with the combined advantages of both convolutional layers and MLP blocks.
Specifically, similar to the MLP Block, the first convolutional layer expands the feature dimension by a ratio, while the second layer maps it back to the original dimension. 
This design does not significantly increase the number of parameters since the feature dimension in this module is typically small, such as $64, 128$. 
Moreover, we argue that, while maintaining parameter efficiency, maximizing the depth of the network and the number of convolution layers is desirable. 
By default, we utilize a patch size of $16$ to process ImageNet with a resolution of $224$. This implies that the SCS requires the stacking of four blocks. Specifically, given an image with dimensions $H\times W$, the four SCS blocks yield spatial resolutions of $\frac{H}{2}\times\frac{W}{2}$, $\frac{H}{4}\times\frac{W}{4}$, $\frac{H}{8}\times\frac{W}{8}$, and $\frac{H}{16}\times\frac{W}{16}$, respectively.

Inspired by ANNs \cite{yuan2021tokens,zhang2023hivit}, the \newoursabbr with SCS adopts a deeper and narrower design compared to its previous version with SPS. Specifically, it has more convolutional layers in the SCS module, resulting in a deeper overall network. 
With a similar number of parameters, the \newoursabbr-8-384 model has lower dimensions compared to the \oursabbrv-8-512 model, indicating a narrower network design.
Table \ref{tab:supervised_learning_imagenet} demonstrates that, under comparable parameter settings, incorporating the SCS into \newoursabbr leads to improved performance in supervised training compared to the original \oursabbrv.

\subsection{Self-supervised Pre-training}\label{sec:spikmae}
\begin{figure}[]
    \centering
    \includegraphics[width=0.9\linewidth]{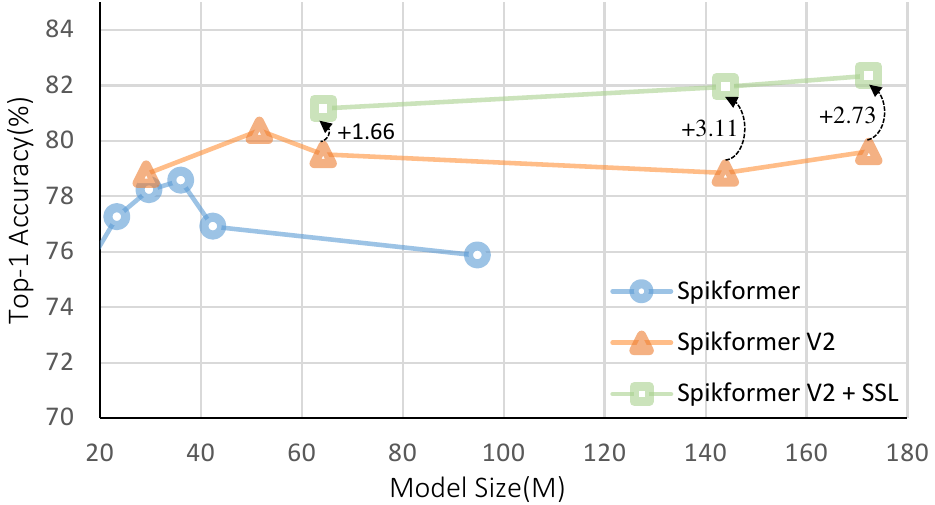}
    \caption{Performance curve with expanded model parameters.
The blue circles represent the performance changes of the previous version of \oursabbrv with increasing network depth and model parameters during supervised training. The orange triangles represent \newoursabbr. These curves indicate that, under supervised learning, the performance of the model tends to decline as the depth and number of parameters increase. In contrast, the green squares represent Spikformer V2 with mask autoencoder pre-training, demonstrating stable performance gains on larger and deeper models.}
    \label{fig:deep}
\end{figure}

Despite the performance improvement achieved by the \newoursabbr, we observe a certain degree of performance degradation when further scaling up and deepening the \newoursabbr, as shown in Fig. \ref{fig:deep}. This phenomenon aligns with what is observed in ANNs \cite{dosovitskiy2020image,touvron2021deit}, mainly attributed to the instability and overfitting during the training of large models.
Concretely, large models such as ViT-Large and ViT-Huge exhibited unstable training and lower performance.
Unlike ANN-Transformers, where the performance saturation or degradation typically occurs above 24 layers \cite{zhou2021deepvit}, the issue becomes apparent in \newoursabbr with 12 or more layers.
The Mask Auto-encoding mechanism \cite{he2022masked} in Self-Supervised Learning can effectively train deep and large models and additionally improve model performance.

In addition to structural improvements in \newoursabbr, we also explore Self-Supervised Learning (SSL) for SNN-Transformers to further enlarge the model and enhance its representational capacity.
SSL has gained significant popularity in ANNs and is considered a fundamental methodology for studying multimodal learning \cite{pmlr-v139-radford21a}, and large-scale models \cite{kirillov2023segany}.
However, the effectiveness of SSL in SNNs has not been extensively investigated. 
It is essential to conduct corresponding research to validate the feasibility and efficacy of SSL in SNNs. 
As a prevalent learning approach observed in the brain \cite{bakhtiari2021functional,millet2022toward}, Self-Supervised Learning can also enhance the representational capacity of neural networks.
Among various Self-Supervised Learning, the Mask Auto-encoding is most suitable for \newoursabbr. 
It adopts a decoupled encoder-decoder architecture to enhance the representation ability of the encoder by reconstructing the masked images using the decoder.

We devise an asymmetric SNN-ANN heterogeneous encoder-decoder architecture, as illustrated in Fig. \ref{fig:spikmae}. The SNN-encoder exclusively processes the partially  observed signal (without mask tokens), while a lightweight ANN-decoder reconstructs the complete signal using the latent representation and mask tokens.
The framework introduces a novel approach that couples SNN and ANN, utilizing the powerful modeling capabilities of ANN to enhance the representational capacity of SNN. 
This represents the third method, following the techniques of ANN2SNN and ANN distillation into SNN, for enhancing SNN through the utilization of ANN.

\textbf{Masking.} Following MAE\cite{he2022masked}, we employ a random sampling strategy for the masking operation on the $N$ patches with a size of $\frac{H}{16}\times\frac{W}{16}$ after SCS splitting. Specifically, we select a subset $N_v$ of patches following a uniform distribution and remove (i.e., mask) the remaining patches $N_m$. Consequently, a $\frac{H}{16}\times\frac{W}{16}$ mask $M$ is generated, where a value of 1 represents the preservation of the corresponding patch in the feature map, while a value of 0 indicates its removal. Despite the fact that the \newoursabbr patches are in the form of spikes, we observed that a high masking rate (e.g., 75\%) remains applicable. This suggests that even in sparse spike signals, redundant information exists, providing insights for the future exploration of efficient SNN training.

Regarding the SCS, we adopt a similar methodology as MC-MAE\cite{gao2022mcmae} to obtain the corresponding masks. Specifically, we upsample the generated mask $M$ by factors of 8, 4, and 2, respectively, to serve as masks for the first, second, and third SCS blocks. The fourth SCS block can directly utilize the mask $M$.

\textbf{\newoursabbr encoder.} The encoder utilized the \newoursabbr with SCS, as described in Sec. \ref{section:SCS}. During the pre-training process, the encoder exclusively processes the visible patches. Apart from utilizing a high masking rate, we further enhance the efficiency of pre-training in Spikformer by setting the time step to 1. This ensures that the reconstruction during pre-training is sufficiently challenging, which is beneficial for the downstream fine-tuning task \cite{he2022masked, gao2022mcmae, tian2023designing}.

\textbf{Transformer decoder.} We employ a lightweight ANN-Transformer as the decoder, which has better image reconstruction ability than the SNN-decoder. The decoder is discarded after pre-training; it does not affect the fact that our \newoursabbr is a pure SNN. As shown in Fig. \ref{fig:spikmae}, the decoder receives a complete set of tokens, comprising encoded visible patches and mask tokens, as input. The mask tokens are replicated from a shared learnable vector, serving as an indicator for the presence of a missing patch that needs to be predicted. Positional embeddings are incorporated into all tokens within the complete set, ensuring that mask tokens possess information about their spatial location in the image.

\textbf{Training Cost.} The motivation is to explore Self-Supervised Learning in SNNs and to consistently enhance the performance of large-scale \newoursabbr. Additionally, we have observed its advantage in reducing training costs. We compare the training cost between supervised training and fine-tuning after pre-training and find it can achieve training acceleration while maintaining stable training and performance improvements, which are shown in Table \ref{tab:spikmae}. As the model parameter size increases, the training acceleration ratio of SSL compared to direct supervised training becomes more prominent. It is worth noting that while in ANNs, pre-training typically involves more than 800 epochs, we find that even with fewer pre-training epochs in SNNs, we can still achieve good fine-tuning results on ImageNet. 
And the finetune performance increases with the increase of pre-training epoches, as shown in Table \ref{tab:ab_spikmae}.

\section{Experiments} \label{sec:5}

In this section, we begin by evaluating the performance of the proposed \oursabbrv and \newoursabbr on the large-scale image classification dataset, ImageNet\cite{deng2009imagenet}. We further conduct an analysis of the structural enhancements incorporated in \newoursabbr, as well as the performance improvements achieved through SSL pre-training. Additionally, we present the performance of \oursabbrv on small-scale static datasets. Furthermore, we evaluate the classification performance of \oursabbrv on popular neuromorphic datasets. Finally, we include ablation experiments and partial visualizations as the concluding part of this section.

\begin{table*}[t]
\small
\caption{Evaluation on ImageNet. Param refers to the number of parameters. 
Power denotes the average theoretical energy consumption during image prediction on the ImageNet test set. \oursabbrv-$L$-$D$ is a \oursabbrv model with $L$ \oursabbrv encoder blocks and $D$ feature embedding dimensions. 
OPs refers to SOPs in SNNs and FLOPs in ANNs.}
\begin{center}
\resizebox{0.9\textwidth}{!}{
\begin{tabular}{cccccccc}
\toprule
\multicolumn{1}{c}{\bf Category} 
&\multicolumn{1}{c}{\bf Methods} 
&\multicolumn{1}{c}{\bf Architecture}
&\multicolumn{1}{c}{\bf \tabincell{c}{Param\\ (M)}}
&\multicolumn{1}{c}{\bf \tabincell{c}{OPs\\ (G)}}
&\multicolumn{1}{c}{\bf \tabincell{c}{Power\\ (mJ)}}
&\multicolumn{1}{c}{\bf\tabincell{c}{Time\\Step}}
&\multicolumn{1}{c}{\bf Acc}\\

 \midrule
 \multicolumn{1}{c}{\multirow{6}{*}{{ANN-to-SNN}}}
    &Hybrid training\cite{rathi2020enabling} &ResNet-34 &21.79 &- &- &250 &61.48\\
    & \multicolumn{1}{c}{\multirow{2}{*}{Spiking ResNet\cite{hu2018residual}}} 
    &ResNet-34 &21.79 &65.28 & {59.295} & 350 & 71.61 \\
    & &ResNet-50 &25.56&78.29 & {70.934} & 350 & 72.75 \\
    &QCFS\cite{bu2022optimal} &VGG-16 &138.42 &- &- &64 &72.85\\
    &Fast-SNN\cite{hu2023fast} &VGG-16 &138.42 &- &44.850 &7 &72.95\\
    &COS\cite{hao2023bridging} &ResNet-34 &21.79 &- &8.070 &8 &74.17\\
    
 \midrule

 \multicolumn{1}{c}{\multirow{5}{*}{{ANN}}} &
   {T2T-ViT\cite{yuan2021tokens}}   & {T2T-ViT-24} & {64.10} & {13.80} & {63.516} & {1} & {\textbf{82.30}}\\
    &{PVT\cite{wang2021pyramid}}   & {PVT-Large} & {61.40} & {9.80} & {38.340} & {1} & {\textbf{81.70}} \\
    &{Swin Transformer\cite{liu2021swin}}   & {SWIN-Base} & {88.00} & {15.40} & {70.880} & {1} & {\textbf{83.50}} \\
    &\multicolumn{1}{c}{\multirow{2}{*}{{MAE\cite{he2022masked}}}}  & {ViT-Base} & {86.60} & {17.60} & {95.310} & {1} & {\textbf{82.30}} \\
    && {ViT-Large} & {307.00} & {-} & {-} & {1} & {\textbf{82.60}} \\
    \midrule
 \multicolumn{1}{c}{\multirow{10}{*}{{Directly Learning}}}
    &\multicolumn{1}{c}{\multirow{2}{*}{TET\cite{deng2021temporal}}} 
   & Spiking-ResNet-34 &21.79 &- &- &6 & {64.79} \\
   & & SEW-ResNet-34 &21.79 &- &- &4 & {68.00} \\
   & \multicolumn{1}{c}{\multirow{1}{*}{STBP-tdBN\cite{zheng2021going}}} &Spiking-ResNet-34 &21.79 &6.50 &{6.393} & 6 & 63.72 \\
    % &Spiking-ResNet-50 &25.56 & & 6 & 63.72 \\
   & \multirow{4}{*}{SEW ResNet\cite{fang2021deep}}  
    & SEW-ResNet-34 &21.79 &3.88 &{4.035} &4 & 67.04 \\
   & & SEW-ResNet-50 &25.56  &4.83 &{4.890} &4 & 67.78 \\
  &  & SEW-ResNet-101 &44.55  &9.30 &{8.913} &4 & 68.76 \\
   & & SEW-ResNet-152 &60.19  &13.72 &{12.891} &4 & 69.26 \\
   &Attention-SNN\cite{yao2023attention} &ResNet-104 &78.37 &- &7.300 &4 &77.08\\
&Spike-driven Transfromer\cite{yao2023spike} &Spiking Transformer-8-768 &66.34 &- &6.090 &4 &77.07\\
&Spikingformer\cite{zhou2023spikingformer} &Spiking Transformer-8-768 &66.34 &12.54 &13.680 &4 &75.85\\
&CML\cite{zhou2023enhancing} &Spiking Transformer-8-768 &66.34 &- &- &4 &77.64\\
 \midrule
  \multicolumn{1}{c}{\multirow{5}{*}{{Directly Learning}}} &
   \multicolumn{1}{c}{\multirow{5}{*}{\textbf{\oursabbrv}}}  &\multicolumn{1}{c}{\multirow{1}{*}{\oursabbrv-8-384}} &16.81&6.63 &5.967 &4 & 75.24\\
   & &\multicolumn{1}{c}{\multirow{1}{*}{\oursabbrv-6-512}}&23.37&8.21 &{7.389} & 4 & 77.26\\
   & &\multicolumn{1}{c}{\multirow{1}{*}{\oursabbrv-8-512}}&29.68&10.95&{9.855} & 4 & \textbf{78.23}\\
   & &\multicolumn{1}{c}{\multirow{1}{*}{\oursabbrv-10-512}}&36.01&13.67 &{12.303} & 4 & \textbf{78.57}\\
    % &\multicolumn{1}{c}{\multirow{1}{*}{\oursabbrv-12-512}}&42.35 & & & 4 & 72.50\\
   & &\multicolumn{1}{c}{\multirow{1}{*}{\oursabbrv-8-768}} &66.34&22.22 &{19.998} & 4 & \textbf{79.55}\\
    \midrule
  \multicolumn{1}{c}{\multirow{4}{*}{{Directly Learning}}} &
   \multicolumn{1}{c}{\multirow{4}{*}{\textbf{\newoursabbr}}}  &\multicolumn{1}{c}{\multirow{2}{*}{\newoursabbr-8-384}} &29.11&1.92 &{1.728} &1 & 75.42\\
    &&&29.11&5.21 &{4.689} &4 & \textbf{78.80}\\
   & &\multicolumn{1}{c}{\multirow{2}{*}{\newoursabbr-8-512}}
    &51.55&3.15&{2.835} & 1 & \textbf{79.05}\\
   &&&51.55&10.40&{9.360} & 4 & \textbf{80.38}\\
    \hline

\end{tabular}}
\end{center}
\label{tab:supervised_learning_imagenet}
\end{table*}

\subsection{Supervised learning experiments on ImageNet} %
\subsubsection{Experimental Settings.}
ImageNet-1K\cite{deng2009imagenet} is a widely used static image dataset for image classification. It consists of 1,000 categories and contains approximately 1.28 million training images and 50,000 test images. 
The images are resized to a default size of 224 × 224 for both training and evaluation.
We employ the AdamW optimizer~\cite{DBLP:journals/corr/KingmaB14} for training, utilizing a cosine decay learning rate scheduler for 300 epochs. Additionally, we incorporate 10 epochs of linear warm-up. Similar to the supervised training strategy in MAE\cite{he2022masked}, we employed a per-iteration learning rate update policy.
We adopt a batch size of 512, distributed across 8 Nvidia V100 GPUs. We set the initial learning rate to 0.0005 and apply a weight decay of 0.05.
In contrast to ANN-Transformers\cite{dosovitskiy2020image,yuan2021tokens}, we do not employ drop path in our approach, as we have found that it does not enhance performance.
Moreover, following DeiT\cite{touvron2021deit,liu2021swin}, data augmentation techniques including RandAugment~\cite{DBLP:conf/nips/CubukZS020}, random erasing~\cite{DBLP:conf/aaai/Zhong0KL020}, and stochastic depth~\cite{DBLP:conf/eccv/HuangSLSW16} are employed for data augmentation in our approach.
Nevertheless, we refrain from incorporating Mixup~\cite{DBLP:conf/iclr/ZhangCDL18} and Cutmix~\cite{DBLP:conf/iccv/YunHCOYC19} for data augmentation in our approach. This decision stems from the recognition that the transformation of static RGB images into spiking event data already poses significant challenges for the classification task. The inclusion of Mixup and Cutmix techniques on spiking data would introduce additional complexities to the classification process on large-scale datasets and potentially impede convergence.

\subsubsection{Results.} 
The supervised learning results of \oursabbrv and \newoursabbr are presented in Table \ref{tab:supervised_learning_imagenet}.
The experimental results demonstrate the superior performance of our proposed \oursabbrv and \newoursabbr compared to previous state-of-the-art models, including both the ANN2SNN and direct training methods, across various model sizes.
Specifically, considering the \newoursabbr-8-512 model with 51.55 million parameters as a representative example, we observe that even when tested on a resolution of 224 with 4 time steps, it achieves an impressive top-1 accuracy of 80.38\% on the ImageNet. 
Compared to the state-of-the-art Attention-SNN \cite{yao2023attention} model, our model achieves a 3.3\% increase in classification accuracy while having half the number of parameters. Note that the A-SNN model is trained using images of resolution 288.
\begin{table*}[t]
\centering
\caption{Comparisons between directly supervised learning and Self-Supervised Learning + finetune. Speedup ratio refers to the supervised learning training time of the \newoursabbr model with the same parameter divided by the SSL training time.}
\setlength{\tabcolsep}{1mm}{
\begin{tabular}{lccccccccc}
    \toprule

    \multicolumn{1}{c}{ \tabincell{c}{Learning Method}} & Model 
    & \multicolumn{1}{c}{ \tabincell{c}{Param\\ (M)}}  & \multicolumn{1}{c}{ \tabincell{c}{Training\\ Epoch}} & \multicolumn{1}{c}{ \tabincell{c}{Training\\ Time (h)}} & \multicolumn{1}{c}{ \tabincell{c}{Speedup}} &Time Step& \multicolumn{1}{c}{ \tabincell{c}{OPs\\ (G)}}&\multicolumn{1}{c}{\tabincell{c}{Power\\ (mJ)}}& Acc. \\
        \midrule
     \multicolumn{1}{c}{\multirow{4}{*}{Learning on ANN}} 
    & {ViT-Base} & {86.60} &300&-& -& {1} & {17.60} & {95.310}  & {\textbf{82.30}}\\
    & {ViT-Large} & {307.00} &300&-& -& {1} & - & -  & {\textbf{82.60}}\\
   & {T2T-ViT-24} & {64.10} &300&-& -& {1} & {13.80} & {63.516}  & {\textbf{82.30}}\\
    & {PVT-Large} & {61.40} &300&-&- & {1} & {9.80} & {38.340}  & {\textbf{81.70}}\\
    & {SWIN-Base} & {88.00} &300&-&-&1& {15.40} & {70.880} & \textbf{83.50}\\

    \midrule
    Supervised Learning& \oursabbrv V1-8-512 & 29.69 &300 & -&-& 4&10.95 &11.304& 78.23 \\
    \midrule
    Pre-training + Finetune & \oursabbrv V1-8-512 & 29.69 &200 + 150 & -&-& 4&- &-& 75.23 \\
    \midrule
  \multicolumn{1}{c}{\multirow{6}{*}{{Supervised Learning}}} &
   \multicolumn{1}{c}{\multirow{2}{*}{\textbf{\newoursabbr-8-512}}} 
    &\multicolumn{1}{c}{\multirow{2}{*}{51.55}} & 300 & 85 & $1 \times$ &1
    &3.15 &{2.835}  & 79.05\\
    & & & -& 102 & - &4
    &10.40 &{9.360}  & 80.38\\
    
    &\multicolumn{1}{c}{\multirow{1}{*}{\textbf{\newoursabbr-12-512}}} &\multicolumn{1}{c}{\multirow{1}{*}{64.18}} & 300 & 100 & $1 \times$ &1
    &4.15 &3.735  & 78.65\\
    
    &\multicolumn{1}{c}{\multirow{1}{*}{\textbf{\newoursabbr-12-768}}} &\multicolumn{1}{c}{\multirow{1}{*}{143.89}} & 300 & 155 & $1 \times$ &1
    &6.35&5.715 & 76.61\\
    &\multicolumn{1}{c}{\multirow{1}{*}{\textbf{\newoursabbr-16-768}}} &\multicolumn{1}{c}{\multirow{1}{*}{172.28}} & 300 & 240 & $1 \times$ &1
    &7.64&6.876 & 77.43\\

    \midrule
    \multicolumn{1}{c}{\multirow{6}{*}{{Pre-training + Finetune}}} 
    &\multicolumn{1}{c}{\multirow{2}{*}{\textbf{\newoursabbr-12-512}}} 
    &\multicolumn{1}{c}{\multirow{2}{*}{64.18}} & 200+150 & 96.7 & $1.03 \times$ &1
    &4.01 &{3.609}  & 79.64\\
    &&   & - & - & $-$ &4
    &16.31&14.680  & \textbf{81.17}\\
    % & & & -& 102 & $0.83\times$ &4
    % &10.40 &{9.491}  & 80.38\\
    &\multicolumn{1}{c}{\multirow{2}{*}{\textbf{\newoursabbr-12-768}}} 
    &\multicolumn{1}{c}{\multirow{2}{*}{143.89}} 
    & 200+150 & 147.5 & $1.05 \times$ &1
    &6.40 &{5.760}  & \textbf{80.49}\\
     &&   & - & - & $-$ &4
    &19.97&{17.970}  & \textbf{81.94}\\

    &\multicolumn{1}{c}{\multirow{2}{*}{\textbf{\newoursabbr-16-768}}} 
    &\multicolumn{1}{c}{\multirow{2}{*}{172.70}} 
    & 200+150 & 196.7 & $1.22 \times$ &1
    &7.48 &{6.732}  & \textbf{81.10}\\
     &&   & - & - & $-$ &4
    &28.42&{25.578}  & \textbf{82.35}\\

    \bottomrule
\end{tabular}
% }
\label{tab:spikmae}}
\end{table*}
To the best of our knowledge, this is the first time that an SNN model has achieved an accuracy of over 80\% on ImageNet, and it is accomplished using the \newoursabbr-base model with low simulation latency (4 steps). This breakthrough challenges the common perception of low performance in SNNs and demonstrates the potential of SNNs.
While achieving high accuracy, \newoursabbr also demonstrates competitive theoretical energy efficiency. 
In comparison to the previous version of Spikformer, the \newoursabbr achieves certain performance improvements (approximately 0.5\%) while maintaining a similar number of parameters.
For example, $29.11M$~\newoursabbr-8-384 achieves a 0.57\% higher performance compared to $29.68M$ \oursabbrv-8-512, while energy consumption is only 0.48 times that of the other.
This demonstrates the effectiveness of SCS and the narrow and deep design. 
The \newoursabbr-8-512, which has fewer parameter ($51.55 M$ VS. $66.34M$), achieves higher classification performance than \newoursabbr-8-768 ($80.38\% $ VS. $79.55\%$). 
Specifically, compared to ANNs, \newoursabbr-8-512 achieves similar accuracy ($80.38\%$) with only one-tenth of the energy consumption.

\newoursabbr also demonstrates excellent performance in a pure binary spiking residual connection, as shown in Table \ref{tab:as}. We replace the ADD operator in the original network with the IAND operator from SEW ResNet\cite{fang2021deep}, transforming the entire network into one that utilizes pure binary activation for inter-layer information propagation. Even in this binary spiking setting, \newoursabbr-8-512 still achieves a performance of $76.12\%$.

Although there is still a performance gap between \oursabbrv and \newoursabbr compared to ANN-Transformers, as shown in Table \ref{tab:supervised_learning_imagenet}, it is not doubtful that our work has taken the first step in exploring high-performance SNN models. We firmly believe that future SNN models can achieve both high performance and low energy consumption.

\subsection{Self-Supervised Learning on ImageNet.}
In order to further enhance the performance of \newoursabbr and address the issue of limited scalability of its model size, we introduce self-supervised pre-training using masked image modeling into \newoursabbr.
\subsubsection{Experimental Settings.}
We conducted SSL pre-training on the ImageNet training set.
The mask ratio is fixed at 75\% based on the original MAE\cite{he2022masked}. The ANN-Transformer decoder consists of 4 transformer layers with 256 feature dimensions and 4 attention heads. During pre-training, we employ a cosine learning rate schedule with 200 epochs, where the first 20 epochs are dedicated to warm-up.  
The AdamW optimizer is adopted with a base learning rate of $1 \times10^{-3}$, a weight decay of 0.05, and a batch size of 1024. 
To augment the data during pre-training, we employed random cropping.
The reconstruction target is the normalized input image. Although the reconstruction results of \oursabbrv as an encoder are inferior to that of MAE in ANN, which is shown in Fig. \ref{fig:img_recon}, fine-tuning performance after pre-training reveals that the RGB image reconstruction task is also applicable to \newoursabbr.

After the pre-training, we perform supervised fine-tuning on the ImageNet training set using the cosine learning rate schedule for 150 epochs. 
We utilize a per-iteration-update cosine decay learning rate strategy with a base learning rate of $1\times 10^{-3}$.
We employ a layer-wise learning rate decay strategy following MAE, where the learning rate of each upper layer in the network is set to 0.7 times the learning rate of the subsequent layer.
To reduce training overhead, we finetune the model using a single-time step model to obtain results on multiple time steps (e.g., 2, 4, 6). The base learning rate is set to $1 \times10^{-5}$, and the fine-tuning is performed for 20 epochs.

\subsubsection{Results.} 

As shown in Table \ref{tab:spikmae}, after pre-training, the performance of \oursabbrv V1 actually decreases, which could be attributed to the presence of max-pooling in the SPS module of \oursabbrv V1. This max-pooling operation leads to information loss during feature extraction and is not suitable for complex image reconstruction tasks during pre-training.

The performance of \newoursabbr is significantly improved through self-supervised pre-training. Moreover, it demonstrates stable performance improvements, particularly in large models.
Specifically, the \newoursabbr-12-768 model achieves an accuracy of 80.49\% using only a single time step. Using one time step, the \newoursabbr-16-768 model achieves an accuracy of 81.10\%, which is 3.67\% higher than the result obtained from supervised directly training (77.43\%).
Although the accuracy is steadily improved, the energy consumption of the model after SSL does not increase significantly, but is similar to that of the supervised training model.
The utilization of SSL accelerates network training, with the acceleration ratio increasing as the network parameters grow in size.
The $172 M$ \newoursabbr-16-768 not only achieves performance enhancement but also experiences a training acceleration of 1.22 times.

\begin{table}%[t]
\caption{Comparison of the performance between \oursabbrv and existing approaches on CIFAR10/100. * denotes self-implementation results by \cite{deng2021temporal}. Note that Hybrid training \cite{rathi2020enabling} adopts ResNet-20 for CIFAR10 and VGG-11 for CIFAR100.}
\begin{adjustbox}{max width=\linewidth} 
\begin{tabular}{ccccccc}
\toprule
  \multicolumn{1}{c}{\bf Methods} &\multicolumn{1}{c}{\bf \tabincell{c}{Param\\ (M)}}
&\bf\tabincell{c}{Time\\Step} &\bf\tabincell{c}{CIFAR10\\Acc} &\bf\tabincell{c}{CIFAR100\\Acc}\\
\midrule
    Hybrid training\cite{rathi2020enabling}  &9.27 &125 &92.22 &67.87\\
    Diet-SNN\cite{rathi2020diet}  &0.27 &10\textbf{/}5  & 92.54& 64.07\\
    STBP\cite{wu2018spatio}  &17.54&12 & 89.83&-\\
    STBP NeuNorm\cite{wu2019direct}  &17.54 &12 &90.53& -\\
    TSSL-BP\cite{zhang2020temporal}  &17.54 &5 &91.41& -\\
    \multicolumn{1}{c}{\multirow{1}{*}{{STBP-tdBN\cite{zheng2021going}}}} &12.63 & 4 & 92.92 & 70.86\\
    TET\cite{deng2021temporal}   &12.63 & 4 & \textbf{94.44}& \textbf{74.47}\\
\midrule
     ResNet-19* &12.63 &1 &\textbf{94.97} & \textbf{75.35}\\
     Transformer-4-384 & {9.32} &1 &{\textbf{96.73}} & {\textbf{81.02}} \\
\midrule
    \multicolumn{1}{c}{\multirow{1}{*}{\oursabbrv-4-256}}&4.15 & 4 & 93.94 & \textbf{75.96}\\
    \multicolumn{1}{c}{\multirow{1}{*}{\oursabbrv-2-384}}&5.76 & 4 & \textbf{94.80} & \textbf{76.95}\\
    \multicolumn{1}{c}{\multirow{1}{*}{\oursabbrv-4-384}}&9.32 & 4 & \textbf{95.19} & \textbf{77.86}\\
    \multicolumn{1}{c}{\multirow{1}{*}{\oursabbrv-4-384 400E}}&9.32 & 4 & \textbf{95.51} & \textbf{78.21}\\
\bottomrule
\end{tabular}
\end{adjustbox}
\label{tab:sd}
\end{table}
\subsection{Small datasets classification}\label{sec:sdc}
We evaluate the performance of \oursabbrv on small-scale datasets, including CIFAR, DVS-Gesture, and CIFAR10DVS. As \newoursabbr is primarily designed for ImageNet, its performance on these smaller datasets was comparatively lower than that of \oursabbrv V1. Because SPS is specifically designed for small datasets and can achieve better performance than SCS.

\subsubsection{CIFAR.} 
The CIFAR dataset consists of 50,000 training images and 10,000 test images, all of which have a resolution of $32\times32$. For our experiments, we use a batch size of 128. We employ a four-block SPS architecture, where the first two blocks do not include the max-pooling layer. The image is divided into 64 $4\times4$ patches. Table \ref{tab:sd} presents the accuracy of our method (\oursabbrv) in comparison to other models on the CIFAR dataset.
As depicted in Table \ref{tab:sd}, our \oursabbrv-$4$-$384$ achieves an accuracy of $95.19\%$ on the CIFAR10 dataset, surpassing the performance of TET ($94.44\%$) and ResNet-19 ANN ($94.97\%$). Notably, the performance continues to improve with increasing dimensions and blocks.
In particular, \oursabbrv-4-384 demonstrates a $1.25\%$ improvement compared to \oursabbrv-4-256 and a $0.39\%$ improvement compared to \oursabbrv-2-384. Moreover, we observe that increasing the number of training epochs to 400 leads to performance enhancement (\oursabbrv-4-384 400E achieves a $0.32\%$ and $0.35\%$ improvement compared to \oursabbrv-4-384 on CIFAR10 and CIFAR100, respectively). Notably, the performance gain of \oursabbrv is even more significant on complex datasets like CIFAR100. Specifically, \oursabbrv-4-384 ($77.86\%, 9.32 \rm M$) achieves a substantial improvement of $2.51\%$ compared to the ResNet-19 ANN ($75.35\%, 12.63\rm M$) model.
However, the performance of ANN-Transformer still surpasses that of \oursabbrv, achieving a $1.22\%$ and $2.81\%$ higher accuracy than the latter on CIFAR10 and CIFAR100, respectively.

\subsubsection{Neuromorphic datasets classification}\label{sec:ndc}
The DVS128 Gesture dataset is a collection of gesture recognition data, consisting of 11 hand gesture categories performed by 29 individuals under 3 illumination conditions. Similarly, CIFAR10-DVS is a neuromorphic dataset derived from the CIFAR10 static image dataset. It involves shifting image samples to capture their spike events with a DVS camera. CIFAR10-DVS consists of 9,000 training samples and 1,000 test samples.

For the aforementioned datasets with an image size of $128 \times 128$, we employ a four-block SPS architecture. The patch embedding dimension is set to $256$, and the patch size is $16 \times 16$. We utilize a shallow \oursabbrv model, which consists of 2 transformer encoder blocks. The \oursattn module includes 8 heads for DVS128 Gesture and 16 heads for CIFAR10-DVS.
A time step of 10 or 16 is utilized.
The training process involves 200 epochs for DVS128 Gesture and 106 epochs for CIFAR10-DVS. The AdamW optimizer is utilized, and the batch size is set to 16. The learning rate starts at 0.001 and is reduced using a cosine decay schedule. For CIFAR10-DVS, we apply data augmentation techniques as described in \cite{li2022neuromorphic}. Additionally, we employ a learnable parameter to control the scaling factor of the $QK^{\rm{T}}V$ computation.

The classification performance of \oursabbrv and the compared state-of-the-art models on neuromorphic datasets is presented in Table \ref{tab:nd}. It can be observed that our model achieves impressive performance on both datasets while utilizing a compact model size of 2.59M parameters. Specifically, on the DVS128 Gesture dataset, we achieve an accuracy of 98.2\% using 16 time steps, outperforming the SEW-ResNet model (97.9\%). Furthermore, our results are competitive with the TA-SNN model (98.6\%, 60 time steps) \cite{yao2021temporal}, which employs floating-point spikes in the forward propagation process.

On the CIFAR10-DVS dataset, our \oursabbrv model outperforms the state-of-the-art methods in terms of accuracy. Specifically, we achieve a 1.6\% and 3.6\% higher accuracy compared to the DSR, which utilizes binary spikes and employs 10 and 16 time steps, respectively (DSR achieves 77.3\% accuracy). It is worth noting that we do not compare our results with the TET in the table since TET is not an architecture-based method but a loss-based method. TET achieves 83.2\% accuracy by using long epochs (300) and a larger model size of 9.27M parameters using VGGSNN.

\begin{table}[t]
% \footnotesize
  \centering
  \caption{Comparison of the performance with state-of-the-art (SOTA) methods on two neuromorphic datasets. Bold font means the best; $^*$ denotes with Data Augmentation.}
  \begin{adjustbox}{max width=\linewidth} 
  \begin{tabular}{lccccccc}
  \toprule 
  \multirow{3}*{{Method}} &
  \multirow{3}*{{Spikes}} & \multicolumn{2}{c}{{CIFAR10-DVS}} &  \multicolumn{2}{c}{{DVS128}} \\
  \cmidrule(l{2pt}r{2pt}){3-4}\cmidrule(l{2pt}r{2pt}){5-6}\cmidrule(l{2pt}r{2pt}){7-8}
    &{} & {$T$ Step} & {Acc} & {$T$ Step} & {Acc} \\
\midrule
  LIAF-Net~\cite{wu2021liaf} & \xmark& 10 & 70.4 & 60 & 97.6 \\ 
    TA-SNN~\cite{yao2021temporal} & \xmark& 10 & 72.0 &60 & 98.6 \\ 
  \midrule
  Rollout~\cite{kugele2020efficient} & \cmark & 48 & 66.8 & 240 & 97.2 \\ 
  DECOLLE~\cite{kaiser2020synaptic}& \cmark & - & - & 500 & 95.5 \\ 

  tdBN~\cite{zheng2021going} & \cmark& 10 & 67.8 &40 & 96.9 \\ 
  PLIF~\cite{fang2021incorporating} & \cmark& 20 & 74.8 &20 & 97.6 \\ 

  SEW-ResNet~\cite{fang2021deep} & \cmark& 16 & 74.4 &16 & 97.9 \\ 
  Dspike~\cite{li2021differentiable} & \cmark& 10 & 75.4$^*$ & - & - \\ 
  SALT~\cite{kim2021optimizing} & \cmark & 20 & 67.1 & - & - \\ 
  DSR~\cite{meng2022training} &\cmark & 10 & 77.3$^*$ & - & - \\ 
  \midrule
  {\multirow{2}{*}{\textbf{\oursabbrv}}} & \cmark & 10 & 78.9$^*$ & 10 &96.9 \\ 
  &\cmark & 16 & \textbf{80.9}$^*$ & 16 & \textbf{98.3} \\ 
  \bottomrule
  \end{tabular}
\end{adjustbox}
   % \vspace{-4mm}
\label{tab:nd}
\end{table}

\subsection{Ablation Studies}

\begin{table}[t]
\centering
\small
  \centering
  \tabcolsep=0.1cm
  \renewcommand{\arraystretch}{0.32}
  \caption{\small Ablation study results on \oursattn, and time step.
  }
\fontsize{8pt}{12pt}\selectfont
% \tiny
% \caption{}
\begin{tabular}{llcc}
 \toprule
 Datasets& Models &\tabincell{c}{Time\\ Step} &\tabincell{c}{Top1-Acc\\(\%)}  \\
 \midrule
 \multicolumn{1}{c}{\multirow{10}{*}{CIFAR10/100}}
 &\multicolumn{1}{l}{\multirow{6}{*}{\oursabbrv-4-384$_{w\ \rm{SSA}}$}}
 &1 & 93.51/74.36\\
 && 2 & 93.59/76.28\\
 && 4 & 95.19/77.86\\
 && 6 & 95.34/78.61\\
   \cmidrule(lr){2-4}
&\oursabbrv-4-384$_{w\ \rm{VSA}}$&4 &94.97/77.92    \\
  &\oursabbrv-4-384$_{w\ \rm{VSA}~V_{\mathcal{F}}}$&4 &95.17/78.37    \\
  \midrule
 \multirow{28}{*}{\tabincell{c}{ImageNet}}
  &\oursabbrv-8-512$_{w\ \rm{I}}$&4 &\xmark   \\
 &\oursabbrv-8-512$_{w\ \rm{ReLU}}$&4 &\xmark   \\
 &\oursabbrv-8-512$_{w\ \rm{LeakyReLU}}$&4 &\xmark   \\
 &\oursabbrv-8-512$_{w\ \rm{VSA}}$&4 &77.92    \\
  &\oursabbrv-8-512$_{w\ \rm{VSA}~V_{\mathcal{F}}}$&4 &79.03    \\
  \cmidrule(lr){2-4}
  &\multicolumn{1}{l}{\multirow{4}{*}{\oursabbrv-8-512$_{w\ \rm{SSA}}$}}
 &1 &76.01    \\
 &&2 &77.18    \\
 &&4 &78.23   \\
 &&6 &78.57  \\
   \cmidrule(lr){2-4}
  &\multicolumn{1}{l}{\multirow{4}{*}{\newoursabbr-8-512$_{w\ \rm{SSA}}$}}
 &1 &79.05    \\
 &&2 &79.20    \\
 &&4 &80.38   \\
 &&6 &80.73  \\
\cmidrule(lr){2-4}
  &\newoursabbr-8-512-IAND
 &1 &76.12    \\
\bottomrule

\end{tabular}
\label{tab:as}
\end{table}
\subsubsection{Time Step}

The accuracy results for different simulation time steps of the spiking neuron are presented in Table \ref{tab:as}. When the time step is set to 1, our method exhibits a 1.87\% decrease in accuracy compared to the network with a time step of 4 on CIFAR10. However, even with only 1 time step, \oursabbrv-8-512 and \newoursabbr-8-512 achieve a respectable accuracy of 76.01\% and 79.05\%, respectively. These findings demonstrate the robustness of \oursabbrv under low latency conditions, where fewer time steps are utilized.

\subsubsection{Setting in SSL}
Mask ratio, pre-training epochs, and finetuning epochs are crucial configuration parameters of SSL. We conduct ablation experiments on these three parameters separately, as depicted in Table \ref{tab:ab_spikmae}. Extending the number of pre-training and fine-tuning epochs has shown potential for yielding performance improvements to a certain extent. Moreover, altering the mask ratio to a certain degree also influences performance. In consideration of a balanced trade-off between training costs and performance, a mask ratio of 75\% has been adopted. The implication of higher mask ratios suggests a similarity to the floating-point feature \cite{he2022masked}, indicating the presence of significant spatial redundancy within the spiking feature as well.

\begin{table}[t]
  \centering
  % \small
  \setlength\tabcolsep{2.5mm}
  \renewcommand\arraystretch{1}
  \caption{
    Ablation experiments of SSL. We use \newoursabbr-12-512 as the encoder on ImageNet classification under various mask ratios, pre-training epoch and fine-tuning epoch configurations.}
  \label{tab:ab_spikmae}
  \begin{tabular}{cccc} \toprule
  Mask Ratio & \tabincell{c}{Pretrain\\Epoch} & \tabincell{c}{Finetune\\Epoch} & Top-1 Acc.  \\ \midrule
    75\%   & 200 & 150 & 79.64  \\ 
    75\%   & 400 & 150 & 79.71  \\ 
    75\%   & 800 & 150 & 79.90  \\ 
    75\%   & 200 & 200 & 79.93  \\ 
    60\%   & 200 & 150 & 79.40  \\ 
    50\%   & 200 & 150 & 79.73  \\ 
    \bottomrule
  \end{tabular}
\end{table}

\subsubsection{Visualization of Image Reconstruction}
\begin{figure}[]
\centering
\includegraphics[width=\linewidth]{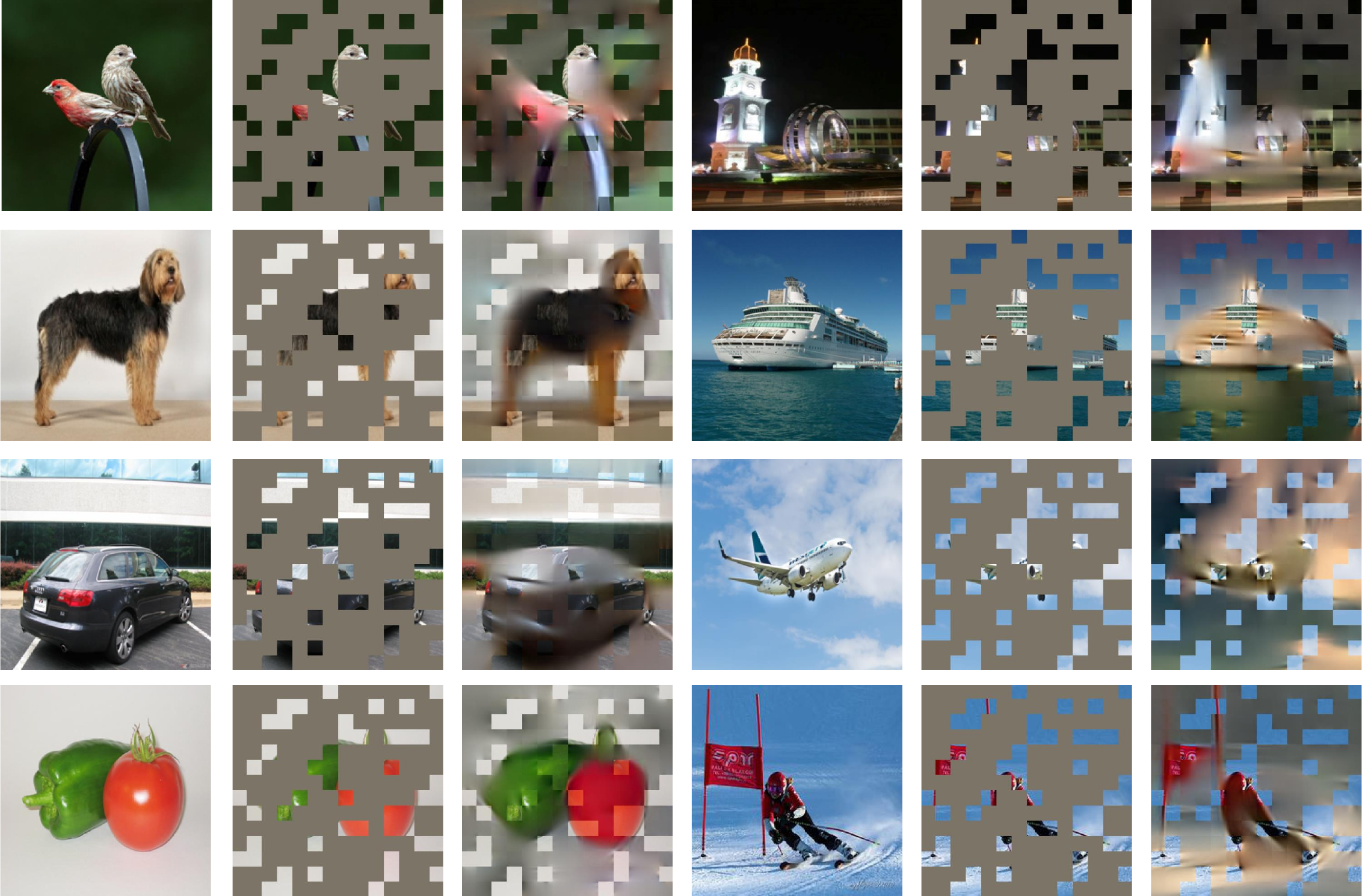}
\caption{Mask and Reconstruction Examples. For each triplet, we present the original image (left), masked image (middle), and reconstructed image (right). The mask ratio is set at 75\%, and the encoder employed is \newoursabbr-12-512.}
\label{fig:img_recon}
\end{figure}

We apply masking to the images and employ the SNN-ANN encoder-decoder framework for reconstruction, as depicted in Fig. \ref{fig:img_recon}. The reconstruction demonstrates favorable outcomes, indicating the effectiveness of pre-training. However, the reconstruction performance is comparatively inferior to that of ANN-MAE. This discrepancy can be attributed to the fact that the task of reconstruction poses challenges for the SNN encoder, which relies on utilizing spiking features for inter-layer propagation. Furthermore, in contrast to the original MAE employing an 8-layer 384-dimensional decoder, we utilize a lightweight 4-layer 256-dimensional ANN decoder. This choice also contributes to the relatively inferior reconstruction performance compared to ANN-MAE.

\section{Conclusions and Future Work} \label{sec:6}
In this paper, we propose the Transformer framework for SNNs, called \oursfull (\oursabbrv), which achieves state-of-the-art performance in SNNs by leveraging the spiking self-attention mechanism.
Besides, to address the issue of low performance of SNNs, including \oursabbrv, on ImageNet, we further optimized the \oursabbrv's architecture and introduced self-supervised pre-training, leading to the proposal of \newoursabbr. Our method achieves a significant breakthrough in the field of SNNs, surpassing the high-accuracy bottleneck on ImageNet (80\%). We hope that the strong performance of \newoursabbr will encourage further research on better SNN-Transformer and SNN-Self-Supervised Learning.

Despite these contributions, certain limitations still exist.
Concretely, experiments on other visual downstream tasks, such as object detection and semantic segmentation, have not been conducted due to the lack of widely adopted detection and segmentation frameworks for direct training of SNNs. 
In the future, we will focus on exploring the performance of Spiking Transformers on various visual tasks and aim to establish a comprehensive visual algorithm system for SNN-Transformers.

\ifCLASSOPTIONcaptionsoff
  \newpage
\fi

\bibliographystyle{IEEEtran}
\bibliography{spikformer-v2}

@article{maass1997networks,
  title={Networks of spiking neurons: the third generation of neural network models},
  author={Maass, Wolfgang},
  journal={Neural networks},
  volume={10},
  number={9},
  pages={1659--1671},
  year={1997},
  publisher={Elsevier}
}

@article{roy2019towards,
  title={Towards spike-based machine intelligence with neuromorphic computing},
  author={Roy, Kaushik and Jaiswal, Akhilesh and Panda, Priyadarshini},
  journal={Nature},
  volume={575},
  number={7784},
  pages={607--617},
  year={2019},
  publisher={Nature Publishing Group UK London}
}

@article{merolla2014million,
  title={A million spiking-neuron integrated circuit with a scalable communication network and interface},
  author={Merolla, Paul A and Arthur, John V and Alvarez-Icaza, Rodrigo and Cassidy, Andrew S and Sawada, Jun and Akopyan, Filipp and Jackson, Bryan L and Imam, Nabil and Guo, Chen and Nakamura, Yutaka and others},
  journal={Science},
  volume={345},
  number={6197},
  pages={668--673},
  year={2014},
  publisher={American Association for the Advancement of Science}
}

@article{davies2018loihi,
  title={Loihi: A neuromorphic manycore processor with on-chip learning},
  author={Davies, Mike and Srinivasa, Narayan and Lin, Tsung-Han and Chinya, Gautham and Cao, Yongqiang and Choday, Sri Harsha and Dimou, Georgios and Joshi, Prasad and Imam, Nabil and Jain, Shweta and others},
  journal={Ieee Micro},
  volume={38},
  number={1},
  pages={82--99},
  year={2018},
  publisher={IEEE}
}

@article{pei2019towards,
  title={Towards artificial general intelligence with hybrid Tianjic chip architecture},
  author={Pei, Jing and Deng, Lei and Song, Sen and Zhao, Mingguo and Zhang, Youhui and Wu, Shuang and Wang, Guanrui and Zou, Zhe and Wu, Zhenzhi and He, Wei and others},
  journal={Nature},
  volume={572},
  number={7767},
  pages={106--111},
  year={2019},
  publisher={Nature Publishing Group UK London}
}

@inproceedings{
dosovitskiy2021an,
title={An Image is Worth 16x16 Words: Transformers for Image Recognition at Scale},
author={Alexey Dosovitskiy and Lucas Beyer and Alexander Kolesnikov and Dirk Weissenborn and Xiaohua Zhai and Thomas Unterthiner and Mostafa Dehghani and Matthias Minderer and Georg Heigold and Sylvain Gelly and Jakob Uszkoreit and Neil Houlsby},
booktitle={International Conference on Learning Representations},
year={2021},
url={https://openreview.net/forum?id=YicbFdNTTy}
}

@inproceedings{touvron2021deit,
  title = {Training data-efficient image transformers \& distillation through attention},
  author =  {Touvron, Hugo and Cord, Matthieu and Douze, Matthijs and Massa, Francisco and Sablayrolles, Alexandre and Jegou, Herve},
  booktitle = {Proceedings of the International Conference on Machine Learning (ICML)},
  pages = 	 {10347--10357},
  year = 	 {2021},
  volume = 	 {139},
}

@inproceedings{yuan2021tokens,
  title={Tokens-to-token vit: Training vision transformers from scratch on imagenet},
  author={Yuan, Li and Chen, Yunpeng and Wang, Tao and Yu, Weihao and Shi, Yujun and Jiang, Zi-Hang and Tay, Francis EH and Feng, Jiashi and Yan, Shuicheng},
  booktitle={Proceedings of the IEEE/CVF International Conference on Computer Vision (ICCV)},
  pages={558--567},
  year={2021}
}

@article{yuan2021volo,
  title={Volo: Vision outlooker for visual recognition},
  author={Yuan, Li and Hou, Qibin and Jiang, Zihang and Feng, Jiashi and Yan, Shuicheng},
  journal={arXiv preprint arXiv:2106.13112},
  year={2021}
}

@inproceedings{wang2021pyramid,
  title={Pyramid vision transformer: A versatile backbone for dense prediction without convolutions},
  author={Wang, Wenhai and Xie, Enze and Li, Xiang and Fan, Deng-Ping and Song, Kaitao and Liang, Ding and Lu, Tong and Luo, Ping and Shao, Ling},
  booktitle={Proceedings of the IEEE/CVF International Conference on Computer Vision (ICCV)},
  pages={568--578},
  year={2021}
}

@inproceedings{liu2021swin,
  title={Swin transformer: Hierarchical vision transformer using shifted windows},
  author={Liu, Ze and Lin, Yutong and Cao, Yue and Hu, Han and Wei, Yixuan and Zhang, Zheng and Lin, Stephen and Guo, Baining},
  booktitle={Proceedings of the IEEE/CVF International Conference on Computer Vision (ICCV)},
  pages={10012--10022},
  year={2021}
}

@inproceedings{chen2021pre,
  title={Pre-trained image processing transformer},
  author={Chen, Hanting and Wang, Yunhe and Guo, Tianyu and Xu, Chang and Deng, Yiping and Liu, Zhenhua and Ma, Siwei and Xu, Chunjing and Xu, Chao and Gao, Wen},
  booktitle={Proceedings of the IEEE/CVF Conference on Computer Vision and Pattern Recognition (CVPR)},
  pages={12299--12310},
  year={2021}
}

@article{chen2023symbolic,
  title={Symbolic discovery of optimization algorithms},
  author={Chen, Xiangning and Liang, Chen and Huang, Da and Real, Esteban and Wang, Kaiyuan and Liu, Yao and Pham, Hieu and Dong, Xuanyi and Luong, Thang and Hsieh, Cho-Jui and others},
  journal={arXiv preprint arXiv:2302.06675},
  year={2023}
}

@inproceedings{fang2021deep,
 title = {{Deep Residual Learning in Spiking Neural Networks}},
 author = {Fang, Wei and Yu, Zhaofei and Chen, Yanqi and Huang, Tiejun and Masquelier, Timoth\'{e}e and Tian, Yonghong},
 booktitle = {Proceedings of the International Conference on Neural Information Processing Systems (NeurIPS)},
 pages = {21056--21069},
 volume = {34},
 year = {2021}
}

@article{hu2021advancing,
  title={Advancing Residual Learning towards Powerful Deep Spiking Neural Networks},
  author={Hu, Yifan and Wu, Yujie and Deng, Lei and Li, Guoqi},
  journal={arXiv preprint arXiv:2112.08954},
  year={2021}
}

@inproceedings{
zhou2023spikformer,
title={Spikformer: When Spiking Neural Network Meets Transformer },
author={Zhaokun Zhou and Yuesheng Zhu and Chao He and Yaowei Wang and Shuicheng YAN and Yonghong Tian and Li Yuan},
booktitle={The Eleventh International Conference on Learning Representations },
year={2023},
url={https://openreview.net/forum?id=frE4fUwz_h}
}

@inproceedings{zheng2021going,
  author    = {Hanle Zheng and
               Yujie Wu and
               Lei Deng and
               Yifan Hu and
               Guoqi Li},
  title     = {{Going Deeper With Directly-Trained Larger Spiking Neural Networks}},
  booktitle = {Proceedings of the AAAI Conference on Artificial Intelligence (AAAI)},
  pages     = {11062--11070},
  year      = {2021}
  
}

@inproceedings{vaswani2017attention,
  title={Attention is all you need},
  author={Vaswani, Ashish and Shazeer, Noam and Parmar, Niki and Uszkoreit, Jakob and Jones, Llion and Gomez, Aidan N and Kaiser, {\L}ukasz and Polosukhin, Illia},
  booktitle={Proceedings of the International Conference on Neural Information Processing Systems (NeurIPS)},
  volume={30},
  year={2017}
}

@article{devlin2018bert,
  title={Bert: Pre-training of deep bidirectional transformers for language understanding},
  author={Devlin, Jacob and Chang, Ming-Wei and Lee, Kenton and Toutanova, Kristina},
  journal={arXiv preprint arXiv:1810.04805},
  year={2018}
}

@article{touvron2023llama,
  title={Llama: Open and efficient foundation language models},
  author={Touvron, Hugo and Lavril, Thibaut and Izacard, Gautier and Martinet, Xavier and Lachaux, Marie-Anne and Lacroix, Timoth{\'e}e and Rozi{\`e}re, Baptiste and Goyal, Naman and Hambro, Eric and Azhar, Faisal and others},
  journal={arXiv preprint arXiv:2302.13971},
  year={2023}
}

@article{brown2020language,
  title={Language models are few-shot learners},
  author={Brown, Tom and Mann, Benjamin and Ryder, Nick and Subbiah, Melanie and Kaplan, Jared D and Dhariwal, Prafulla and Neelakantan, Arvind and Shyam, Pranav and Sastry, Girish and Askell, Amanda and others},
  journal={Advances in neural information processing systems},
  volume={33},
  pages={1877--1901},
  year={2020}
}

@inproceedings{carion2020end,
  title={End-to-end object detection with transformers},
  author={Carion, Nicolas and Massa, Francisco and Synnaeve, Gabriel and Usunier, Nicolas and Kirillov, Alexander and Zagoruyko, Sergey},
  booktitle={Proceedings of the European Conference on Computer Vision (ECCV)},
  pages={213--229},
  year={2020},
  organization={Springer}
}

@article{zhu2020deformable,
  title={Deformable DETR: Deformable Transformers for End-to-End Object Detection},
  author={Zhu, Xizhou and Su, Weijie and Lu, Lewei and Li, Bin and Wang, Xiaogang and Dai, Jifeng},
  journal={arXiv preprint arXiv:2010.04159},
  year={2020}
}

@inproceedings{dosovitskiy2020image,
  title={An image is worth 16x16 words: Transformers for image recognition at scale},
  author={Dosovitskiy, Alexey and Beyer, Lucas and Kolesnikov, Alexander and Weissenborn, Dirk and Zhai, Xiaohua and Unterthiner, Thomas and Dehghani, Mostafa and Minderer, Matthias and Heigold, Georg and Gelly, Sylvain and others},
  booktitle={International Conference on Learning Representa-
tions (ICLR)},
  year={2020}
}

@article{kirillov2023segment,
  title={Segment anything},
  author={Kirillov, Alexander and Mintun, Eric and Ravi, Nikhila and Mao, Hanzi and Rolland, Chloe and Gustafson, Laura and Xiao, Tete and Whitehead, Spencer and Berg, Alexander C and Lo, Wan-Yen and others},
  journal={arXiv preprint arXiv:2304.02643},
  year={2023}
}

@inproceedings{yao2021temporal,
  title={Temporal-wise attention spiking neural networks for event streams classification},
  author={Yao, Man and Gao, Huanhuan and Zhao, Guangshe and Wang, Dingheng and Lin, Yihan and Yang, Zhaoxu and Li, Guoqi},
  booktitle={Proceedings of the IEEE/CVF International Conference on Computer Vision (ICCV)},
  pages={10221--10230},
  year={2021}
}

@article{yao2023attention,
  title={Attention spiking neural networks},
  author={Yao, Man and Zhao, Guangshe and Zhang, Hengyu and Hu, Yifan and Deng, Lei and Tian, Yonghong and Xu, Bo and Li, Guoqi},
  journal={IEEE Transactions on Pattern Analysis and Machine Intelligence},
  year={2023},
  publisher={IEEE}
}

@article{qin2022cosformer,
  title={cosFormer: Rethinking Softmax in Attention},
  author={Qin, Zhen and Sun, Weixuan and Deng, Hui and Li, Dongxu and Wei, Yunshen and Lv, Baohong and Yan, Junjie and Kong, Lingpeng and Zhong, Yiran},
  journal={arXiv preprint arXiv:2202.08791},
  year={2022}
}

@article{choromanski2020rethinking,
  title={Rethinking attention with performers},
  author={Choromanski, Krzysztof and Likhosherstov, Valerii and Dohan, David and Song, Xingyou and Gane, Andreea and Sarlos, Tamas and Hawkins, Peter and Davis, Jared and Mohiuddin, Afroz and Kaiser, Lukasz and others},
  journal={arXiv preprint arXiv:2009.14794},
  year={2020}
}

@article{krizhevsky2009learning,
  title={Learning multiple layers of features from tiny images},
  author={Krizhevsky, Alex},
  journal={},
  year={2009},
  publisher={Citeseer}
}

@inproceedings{deng2009imagenet,
  title={Imagenet: A large-scale hierarchical image database},
  author={Deng, Jia and Dong, Wei and Socher, Richard and Li, Li-Jia and Li, Kai and Fei-Fei, Li},
  booktitle={Proceedings of the IEEE/CVF Conference on Computer Vision and Pattern Recognition (CVPR)},
  pages={248--255},
  year={2009},
}

@article{li2017cifar10,
  title={Cifar10-dvs: an event-stream dataset for object classification},
  author={Li, Hongmin and Liu, Hanchao and Ji, Xiangyang and Li, Guoqi and Shi, Luping},
  journal={Frontiers in neuroscience},
  volume={11},
  pages={309},
  year={2017},
  publisher={Frontiers Media SA}
}

@article{zhuang2021unsupervised,
  title={Unsupervised neural network models of the ventral visual stream},
  author={Zhuang, Chengxu and Yan, Siming and Nayebi, Aran and Schrimpf, Martin and Frank, Michael C and DiCarlo, James J and Yamins, Daniel LK},
  journal={Proceedings of the National Academy of Sciences},
  volume={118},
  number={3},
  pages={e2014196118},
  year={2021},
  publisher={National Acad Sciences}
}

@inproceedings{zhuang2022well,
  title={How Well Do Unsupervised Learning Algorithms Model Human Real-time and Life-long Learning?},
  author={Zhuang, Chengxu and Xiang, Violet and Bai, Yoon and Jia, Xiaoxuan and Turk-Browne, Nicholas and Norman, Kenneth and DiCarlo, James J and Yamins, Daniel LK},
  booktitle={Thirty-sixth Conference on Neural Information Processing Systems Datasets and Benchmarks Track},
  year={2022}
}

@inproceedings{he2022masked,
  title={Masked autoencoders are scalable vision learners},
  author={He, Kaiming and Chen, Xinlei and Xie, Saining and Li, Yanghao and Doll{\'a}r, Piotr and Girshick, Ross},
  booktitle={Proceedings of the IEEE/CVF Conference on Computer Vision and Pattern Recognition},
  pages={16000--16009},
  year={2022}
}

@article{rathi2020diet,
  title={Diet-snn: Direct input encoding with leakage and threshold optimization in deep spiking neural networks},
  author={Rathi, Nitin and Roy, Kaushik},
  journal={arXiv preprint arXiv:2008.03658},
  year={2020}
}

@article{wu2018spatio,
  title={Spatio-temporal backpropagation for training high-performance spiking neural networks},
  author={Wu, Yujie and Deng, Lei and Li, Guoqi and Zhu, Jun and Shi, Luping},
  journal={Frontiers in neuroscience},
  volume={12},
  pages={331},
  year={2018},
  publisher={Frontiers Media SA}
}

@inproceedings{wu2019direct,
  author    = {Yujie Wu and
               Lei Deng and
               Guoqi Li and
               Jun Zhu and
               Yuan Xie and
               Luping Shi},
  title     = {{Direct Training for Spiking Neural Networks: Faster, Larger, Better}},
  booktitle = {Proceedings of the AAAI Conference on Artificial Intelligence (AAAI)},
  pages     = {1311--1318},
  year      = {2019},
  doi       = {10.1609/aaai.v33i01.33011311},
}

@inproceedings{zhang2020temporal,
  title={Temporal spike sequence learning via backpropagation for deep spiking neural networks},
  author={Zhang, Wenrui and Li, Peng},
  booktitle={Proceedings of the International Conference on Neural Information Processing Systems (NeurIPS)},
  volume={33},
  pages={12022--12033},
  year={2020}
}

@inproceedings{
bao2022beit,
title={{BE}iT: {BERT} Pre-Training of Image Transformers},
author={Hangbo Bao and Li Dong and Songhao Piao and Furu Wei},
booktitle={International Conference on Learning Representations},
year={2022},
url={https://openreview.net/forum?id=p-BhZSz59o4}
}

@article{gao2022mcmae,
  title={MCMAE: Masked convolution meets masked autoencoders},
  author={Gao, Peng and Ma, Teli and Li, Hongsheng and Lin, Ziyi and Dai, Jifeng and Qiao, Yu},
  journal={Advances in Neural Information Processing Systems},
  volume={35},
  pages={35632--35644},
  year={2022}
}

@inproceedings{
tian2023designing,
title={Designing {BERT} for Convolutional Networks: Sparse and Hierarchical Masked Modeling},
author={Keyu Tian and Yi Jiang and qishuai diao and Chen Lin and Liwei Wang and Zehuan Yuan},
booktitle={The Eleventh International Conference on Learning Representations },
year={2023},
url={https://openreview.net/forum?id=NRxydtWup1S}
}

@inproceedings{deng2021temporal,
  title={{Temporal Efficient Training of Spiking Neural Network via Gradient Re-weighting}},
  author={Deng, Shikuang and Li, Yuhang and Zhang, Shanghang and Gu, Shi},
  booktitle={International Conference on Learning Representations (ICLR)},
  year={2021}
}

@article{hu2018residual,
  author={Hu, Yangfan and Tang, Huajin and Pan, Gang},
  journal={IEEE Transactions on Neural Networks and Learning Systems}, 
  title={Spiking Deep Residual Networks}, 
  year={2021},
  volume={},
  number={},
  pages={1-6},
  doi={10.1109/TNNLS.2021.3119238}}

@article{rathi2020enabling,
  title={Enabling deep spiking neural networks with hybrid conversion and spike timing dependent backpropagation},
  author={Rathi, Nitin and Srinivasan, Gopalakrishnan and Panda, Priyadarshini and Roy, Kaushik},
  journal={arXiv preprint arXiv:2005.01807},
  year={2020}
}

@inproceedings{
hao2023bridging,
title={Bridging the Gap between {ANN}s and {SNN}s by Calibrating Offset Spikes},
author={Zecheng Hao and Jianhao Ding and Tong Bu and Tiejun Huang and Zhaofei Yu},
booktitle={The Eleventh International Conference on Learning Representations },
year={2023},
url={https://openreview.net/forum?id=PFbzoWZyZRX}
}

@article{hu2023fast,
  title={Fast-SNN: Fast Spiking Neural Network by Converting Quantized ANN},
  author={Hu, Yangfan and Zheng, Qian and Jiang, Xudong and Pan, Gang},
  journal={arXiv preprint arXiv:2305.19868},
  year={2023}
}

@inproceedings{
bu2022optimal,
title={Optimal {ANN}-{SNN} Conversion for High-accuracy and Ultra-low-latency Spiking Neural Networks},
author={Tong Bu and Wei Fang and Jianhao Ding and PENGLIN DAI and Zhaofei Yu and Tiejun Huang},
booktitle={International Conference on Learning Representations},
year={2022},
url={https://openreview.net/forum?id=7B3IJMM1k_M}
}

@inproceedings{NEURIPS2021_61f2585b,
 author = {Perez-Nieves, Nicolas and Goodman, Dan},
 booktitle = {Advances in Neural Information Processing Systems},
 editor = {M. Ranzato and A. Beygelzimer and Y. Dauphin and P.S. Liang and J. Wortman Vaughan},
 pages = {11795--11808},
 publisher = {Curran Associates, Inc.},
 title = {Sparse Spiking Gradient Descent},
 url = {https://proceedings.neurips.cc/paper_files/paper/2021/file/61f2585b0ebcf1f532c4d1ec9a7d51aa-Paper.pdf},
 volume = {34},
 year = {2021}
}

@inproceedings{katharopoulos2020transformers,
  title={Transformers are rnns: Fast autoregressive transformers with linear attention},
  author={Katharopoulos, Angelos and Vyas, Apoorv and Pappas, Nikolaos and Fleuret, Fran{\c{c}}ois},
  booktitle={Proceedings of the 37th International Conference on Machine Learning (ICML)},
  pages={5156--5165},
  year={2020},
}

@article{song2021ufo,
  title={Ufo-vit: High performance linear vision transformer without softmax},
  author={Song, Jeong-geun},
  journal={arXiv preprint arXiv:2109.14382},
  year={2021}
}

@inproceedings{yang2021focal,
  title={Focal attention for long-range interactions in vision transformers},
  author={Yang, Jianwei and Li, Chunyuan and Zhang, Pengchuan and Dai, Xiyang and Xiao, Bin and Yuan, Lu and Gao, Jianfeng},
  booktitle={Proceedings of the International Conference on Neural Information Processing Systems (NeurIPS)},
  volume={34},
  pages={30008--30022},
  year={2021}
}

@inproceedings{rao2021dynamicvit,
  title={Dynamicvit: Efficient vision transformers with dynamic token sparsification},
  author={Rao, Yongming and Zhao, Wenliang and Liu, Benlin and Lu, Jiwen and Zhou, Jie and Hsieh, Cho-Jui},
  booktitle={Proceedings of the International Conference on Neural Information Processing Systems (NeurIPS)},
  volume={34},
  pages={13937--13949},
  year={2021}
}

@inproceedings{chen2020simple,
  title={A simple framework for contrastive learning of visual representations},
  author={Chen, Ting and Kornblith, Simon and Norouzi, Mohammad and Hinton, Geoffrey},
  booktitle={International conference on machine learning},
  pages={1597--1607},
  year={2020},
  organization={PMLR}
}

@inproceedings{he2020momentum,
  title={Momentum contrast for unsupervised visual representation learning},
  author={He, Kaiming and Fan, Haoqi and Wu, Yuxin and Xie, Saining and Girshick, Ross},
  booktitle={Proceedings of the IEEE/CVF conference on computer vision and pattern recognition},
  pages={9729--9738},
  year={2020}
}

@inproceedings{caron2021emerging,
  title={Emerging properties in self-supervised vision transformers},
  author={Caron, Mathilde and Touvron, Hugo and Misra, Ishan and J{\'e}gou, Herv{\'e} and Mairal, Julien and Bojanowski, Piotr and Joulin, Armand},
  booktitle={Proceedings of the IEEE/CVF international conference on computer vision},
  pages={9650--9660},
  year={2021}
}

@InProceedings{Chen_2021_ICCV,
    author    = {Chen, Xinlei and Xie, Saining and He, Kaiming},
    title     = {An Empirical Study of Training Self-Supervised Vision Transformers},
    booktitle = {Proceedings of the IEEE/CVF International Conference on Computer Vision (ICCV)},
    month     = {October},
    year      = {2021},
    pages     = {9640-9649}
}

@article{cao2015spiking,
  title={Spiking deep convolutional neural networks for energy-efficient object recognition},
  author={Cao, Yongqiang and Chen, Yang and Khosla, Deepak},
  journal={International Journal of Computer Vision},
  volume={113},
  number={1},
  pages={54--66},
  year={2015},
  publisher={Springer}
}

@article{hunsberger2015spiking,
  title={Spiking deep networks with LIF neurons},
  author={Hunsberger, Eric and Eliasmith, Chris},
  journal={arXiv preprint arXiv:1510.08829},
  year={2015}
}

@article{rueckauer2017conversion,
  title={Conversion of continuous-valued deep networks to efficient event-driven networks for image classification},
  author={Rueckauer, Bodo and Lungu, Iulia-Alexandra and Hu, Yuhuang and Pfeiffer, Michael and Liu, Shih-Chii},
  journal={Frontiers in neuroscience},
  volume={11},
  pages={682},
  year={2017},
  publisher={Frontiers Media SA}
}

@inproceedings{wang2022signed,
  title={Signed Neuron with Memory: Towards Simple, Accurate and High-Efficient ANN-SNN Conversion},
  author={Wang, Yuchen and Zhang, Malu and Chen, Yi and Qu, Hong},
  booktitle={International Joint Conference on Artificial Intelligence},
  year={2022}
}

@article{meng2022training,
  title={{Training High-Performance Low-Latency Spiking Neural Networks by Differentiation on Spike Representation}},
  author={Meng, Qingyan and Xiao, Mingqing and Yan, Shen and Wang, Yisen and Lin, Zhouchen and Luo, Zhi-Quan},
  journal={ArXiv preprint arXiv:2205.00459},
  year={2022}
}

@inproceedings{han2020rmp,
  title={Rmp-snn: Residual membrane potential neuron for enabling deeper high-accuracy and low-latency spiking neural network},
  author={Han, Bing and Srinivasan, Gopalakrishnan and Roy, Kaushik},
  booktitle={Proceedings of the IEEE/CVF Conference on Computer Vision and Pattern Recognition (CVPR)},
  pages={13558--13567},
  year={2020}
}

@article{DBLP:journals/corr/abs-2102-12122,
	author    = {Wenhai Wang and
	Enze Xie and
	Xiang Li and
	Deng{-}Ping Fan and
	Kaitao Song and
	Ding Liang and
	Tong Lu and
	Ping Luo and
	Ling Shao},
	title     = {Pyramid Vision Transformer: {A} Versatile Backbone for Dense Prediction
	without Convolutions},
	journal   = {CoRR},
	volume    = {abs/2102.12122},
	year      = {2021},
	archivePrefix = {arXiv},
}

@inproceedings{chu2021twins,
  title={Twins: Revisiting the design of spatial attention in vision transformers},
  author={Chu, Xiangxiang and Tian, Zhi and Wang, Yuqing and Zhang, Bo and Ren, Haibing and Wei, Xiaolin and Xia, Huaxia and Shen, Chunhua},
  booktitle={Proceedings of the International Conference on Neural Information Processing Systems (NeurIPS)},
  volume={34},
  pages={9355--9366},
  year={2021}
}

@inproceedings{xiao2021early,
  title={Early convolutions help transformers see better},
  author={Xiao, Tete and Singh, Mannat and Mintun, Eric and Darrell, Trevor and Doll{\'a}r, Piotr and Girshick, Ross},
  booktitle={Proceedings of the International Conference on Neural Information Processing Systems (NeurIPS)},
  volume={34},
  pages={30392--30400},
  year={2021}
}

@article{hassani2021escaping,
  title={Escaping the big data paradigm with compact transformers},
  author={Hassani, Ali and Walton, Steven and Shah, Nikhil and Abuduweili, Abulikemu and Li, Jiachen and Shi, Humphrey},
  journal={arXiv preprint arXiv:2104.05704},
  year={2021}
}

@article{zhou2023enhancing,
  title={Enhancing the Performance of Transformer-based Spiking Neural Networks by Improved Downsampling with Precise Gradient Backpropagation},
  author={Zhou, Chenlin and Zhang, Han and Zhou, Zhaokun and Yu, Liutao and Ma, Zhengyu and Zhou, Huihui and Fan, Xiaopeng and Tian, Yonghong},
  journal={arXiv preprint arXiv:2305.05954},
  year={2023}
}

@inproceedings{zhang2023hivit,
  title={Hivit: A simpler and more efficient design of hierarchical vision transformer},
  author={Zhang, Xiaosong and Tian, Yunjie and Xie, Lingxi and Huang, Wei and Dai, Qi and Ye, Qixiang and Tian, Qi},
  booktitle={The Eleventh International Conference on Learning Representations},
  year={2023}
}

@article{bakhtiari2021functional,
  title={The functional specialization of visual cortex emerges from training parallel pathways with self-supervised predictive learning},
  author={Bakhtiari, Shahab and Mineault, Patrick and Lillicrap, Timothy and Pack, Christopher and Richards, Blake},
  journal={Advances in Neural Information Processing Systems},
  volume={34},
  pages={25164--25178},
  year={2021}
}

@article{millet2022toward,
  title={Toward a realistic model of speech processing in the brain with self-supervised learning},
  author={Millet, Juliette and Caucheteux, Charlotte and Boubenec, Yves and Gramfort, Alexandre and Dunbar, Ewan and Pallier, Christophe and King, Jean-Remi and others},
  journal={Advances in Neural Information Processing Systems},
  volume={35},
  pages={33428--33443},
  year={2022}
}

@inproceedings{li2022neuromorphic,
  title={Neuromorphic data augmentation for training spiking neural networks},
  author={Li, Yuhang and Kim, Youngeun and Park, Hyoungseob and Geller, Tamar and Panda, Priyadarshini},
  booktitle={European Conference on Computer Vision},
  pages={631--649},
  year={2022},
  organization={Springer}
}

@article{kugele2020efficient,
  title={{Efficient Processing of Spatio-temporal Data Streams with Spiking Neural Networks}},
  author={Kugele, Alexander and Pfeil, Thomas and Pfeiffer, Michael and Chicca, Elisabetta},
  journal={Frontiers in Neuroscience},
  volume={14},
  pages={439},
  year={2020}
}

@article{kaiser2020synaptic,
  title={{Synaptic Plasticity Dynamics for Deep Continuous Local Learning (DECOLLE)}},
  author={Kaiser, Jacques and Mostafa, Hesham and Neftci, Emre},
  journal={Frontiers in Neuroscience},
  volume={14},
  pages={424},
  year={2020},
  DOI={10.3389/fnins.2020.00424},      
}

@ARTICLE{wu2021liaf,
  author={Wu, Zhenzhi and Zhang, Hehui and Lin, Yihan and Li, Guoqi and Wang, Meng and Tang, Ye},
  journal={IEEE Transactions on Neural Networks and Learning Systems}, 
  title={{LIAF-Net: Leaky Integrate and Analog Fire Network for Lightweight and Efficient Spatiotemporal Information Processing}}, 
  year={2021},
  volume={},
  number={},
  pages={1-14},
  doi={10.1109/TNNLS.2021.3073016}}

@inproceedings{fang2021incorporating,
  title={Incorporating learnable membrane time constant to enhance learning of spiking neural networks},
  author={Fang, Wei and Yu, Zhaofei and Chen, Yanqi and Masquelier, Timoth{\'e}e and Huang, Tiejun and Tian, Yonghong},
  booktitle={Proceedings of the IEEE/CVF International Conference on Computer Vision (ICCV)},
  pages={2661--2671},
  year={2021}
}

@inproceedings{li2021differentiable,
 author = {Li, Yuhang and Guo, Yufei and Zhang, Shanghang and Deng, Shikuang and Hai, Yongqing and Gu, Shi},
 booktitle = {Proceedings of the International Conference on
Neural Information Processing Systems (NeurIPS)},
 pages = {23426--23439},
 title = {{Differentiable Spike: Rethinking Gradient-Descent for Training Spiking Neural Networks}},
 volume = {34},
 year = {2021}
}

@article{kim2021optimizing,
  title={{Optimizing Deeper Spiking Neural Networks for Dynamic Vision Sensing}},
  author={Kim, Youngeun and Panda, Priyadarshini},
  journal={Neural Networks},
  volume={144},
  pages={686--698},
  year={2021}
}

@article{brohan2022rt,
  title={Rt-1: Robotics transformer for real-world control at scale},
  author={Brohan, Anthony and Brown, Noah and Carbajal, Justice and Chebotar, Yevgen and Dabis, Joseph and Finn, Chelsea and Gopalakrishnan, Keerthana and Hausman, Karol and Herzog, Alex and Hsu, Jasmine and others},
  journal={arXiv preprint arXiv:2212.06817},
  year={2022}
}

@InProceedings{pmlr-v139-radford21a,
  title = 	 {Learning Transferable Visual Models From Natural Language Supervision},
  author =       {Radford, Alec and Kim, Jong Wook and Hallacy, Chris and Ramesh, Aditya and Goh, Gabriel and Agarwal, Sandhini and Sastry, Girish and Askell, Amanda and Mishkin, Pamela and Clark, Jack and Krueger, Gretchen and Sutskever, Ilya},
  booktitle = 	 {Proceedings of the 38th International Conference on Machine Learning},
  pages = 	 {8748--8763},
  year = 	 {2021},
  editor = 	 {Meila, Marina and Zhang, Tong},
  volume = 	 {139},
  series = 	 {Proceedings of Machine Learning Research},
  month = 	 {18--24 Jul},
  publisher =    {PMLR},
}

@article{kirillov2023segany,
  title={Segment Anything},
  author={Kirillov, Alexander and Mintun, Eric and Ravi, Nikhila and Mao, Hanzi and Rolland, Chloe and Gustafson, Laura and Xiao, Tete and Whitehead, Spencer and Berg, Alexander C. and Lo, Wan-Yen and Doll{\'a}r, Piotr and Girshick, Ross},
  journal={arXiv:2304.02643},
  year={2023}
}

@misc{zhou2021deepvit,
      title={DeepViT: Towards Deeper Vision Transformer}, 
      author={Daquan Zhou and Bingyi Kang and Xiaojie Jin and Linjie Yang and Xiaochen Lian and Zihang Jiang and Qibin Hou and Jiashi Feng},
      year={2021},
      eprint={2103.11886},
      archivePrefix={arXiv},
      primaryClass={cs.CV}
}

@article{wu2021progressive,
  title={Progressive tandem learning for pattern recognition with deep spiking neural networks},
  author={Wu, Jibin and Xu, Chenglin and Han, Xiao and Zhou, Daquan and Zhang, Malu and Li, Haizhou and Tan, Kay Chen},
  journal={IEEE Transactions on Pattern Analysis and Machine Intelligence},
  volume={44},
  number={11},
  pages={7824--7840},
  year={2021},
  publisher={IEEE}
}

@inproceedings{kundu2021hire,
  title={Hire-snn: Harnessing the inherent robustness of energy-efficient deep spiking neural networks by training with crafted input noise},
  author={Kundu, Souvik and Pedram, Massoud and Beerel, Peter A},
  booktitle={Proceedings of the IEEE/CVF International Conference on Computer Vision (ICCV)},
  pages={5209--5218},
  year={2021}
}

@article{DBLP:journals/corr/abs-2103-14030,
	author    = {Ze Liu and
	Yutong Lin and
	Yue Cao and
	Han Hu and
	Yixuan Wei and
	Zheng Zhang and
	Stephen Lin and
	Baining Guo},
	title     = {Swin Transformer: Hierarchical Vision Transformer using Shifted Windows},
	journal   = {CoRR},
	volume    = {abs/2103.14030},
	year      = {2021},
	archivePrefix = {arXiv},
}

@article{yao2023spike,
  title={Spike-driven Transformer},
  author={Yao, Man and Hu, Jiakui and Zhou, Zhaokun and Yuan, Li and Tian, Yonghong and Xu, Bo and Li, Guoqi},
  journal={arXiv preprint arXiv:2307.01694},
  year={2023}
}

@article{zhou2023spikingformer,
  title={Spikingformer: Spike-driven Residual Learning for Transformer-based Spiking Neural Network},
  author={Zhou, Chenlin and Yu, Liutao and Zhou, Zhaokun and Zhang, Han and Ma, Zhengyu and Zhou, Huihui and Tian, Yonghong},
  journal={arXiv preprint arXiv:2304.11954},
  year={2023}
}

@inproceedings{DBLP:journals/corr/KingmaB14,
	author    = {Diederik P. Kingma and
	Jimmy Ba},
	title     = {Adam: {A} Method for Stochastic Optimization},
	booktitle = {International Conference on Learning Representations, {ICLR}},
	year      = {2015},
}

@inproceedings{DBLP:conf/nips/CubukZS020,
	author    = {Ekin Dogus Cubuk and
	Barret Zoph and
	Jon Shlens and
	Quoc Le},
	title     = {RandAugment: Practical Automated Data Augmentation with a Reduced
	Search Space},
	booktitle = {Neural Information Processing Systems, {NeurIPS}},
	year      = {2020},
}

@inproceedings{DBLP:conf/iclr/ZhangCDL18,
	author    = {Hongyi Zhang and
	Moustapha Ciss{\'{e}} and
	Yann N. Dauphin and
	David Lopez{-}Paz},
	title     = {mixup: Beyond Empirical Risk Minimization},
	booktitle = {International Conference on Learning Representations, {ICLR}},
	year      = {2018},
}

@inproceedings{DBLP:conf/iccv/YunHCOYC19,
	author    = {Sangdoo Yun and
	Dongyoon Han and
	Sanghyuk Chun and
	Seong Joon Oh and
	Youngjoon Yoo and
	Junsuk Choe},
	title     = {CutMix: Regularization Strategy to Train Strong Classifiers With Localizable
	Features},
	booktitle = {International Conference on Computer Vision, {ICCV}},
	pages     = {6022--6031},
	year      = {2019},
}

@inproceedings{DBLP:conf/aaai/Zhong0KL020,
	author    = {Zhun Zhong and
	Liang Zheng and
	Guoliang Kang and
	Shaozi Li and
	Yi Yang},
	title     = {Random Erasing Data Augmentation},
	booktitle = {Association for the Advancement of Artificial Intelligence, {AAAI}},
	pages     = {13001--13008},
	year      = {2020},
}

@inproceedings{DBLP:conf/eccv/HuangSLSW16,
	author    = {Gao Huang and
	Yu Sun and
	Zhuang Liu and
	Daniel Sedra and
	Kilian Q. Weinberger},
	editor    = {Bastian Leibe and
	Jiri Matas and
	Nicu Sebe and
	Max Welling},
	title     = {Deep Networks with Stochastic Depth},
	booktitle = {European Conference on Computer Vision, {ECCV}},
	volume    = {9908},
	pages     = {646--661},
	year      = {2016},
}

@article{cheng2023meta,
  title={Meta neurons improve spiking neural networks for efficient spatio-temporal learning},
  author={Cheng, Xiang and Zhang, Tielin and Jia, Shuncheng and Xu, Bo},
  journal={Neurocomputing},
  volume={531},
  pages={217--225},
  year={2023},
  publisher={Elsevier}
}

@article{zhu2023spikegpt,
  title={Spikegpt: Generative pre-trained language model with spiking neural networks},
  author={Zhu, Rui-Jie and Zhao, Qihang and Eshraghian, Jason K},
  journal={arXiv preprint arXiv:2302.13939},
  year={2023}
}

@article{Li2021MViTv2IM,
  title={MViTv2: Improved Multiscale Vision Transformers for Classification and Detection},
  author={Yanghao Li and Chaoxia Wu and Haoqi Fan and Karttikeya Mangalam and Bo Xiong and Jitendra Malik and Christoph Feichtenhofer},
  journal={2022 IEEE/CVF Conference on Computer Vision and Pattern Recognition (CVPR)},
  year={2021},
  pages={4794-4804}
}

@article{lee2016training,
  title={Training deep spiking neural networks using backpropagation},
  author={Lee, Jun Haeng and Delbruck, Tobi and Pfeiffer, Michael},
  journal={Frontiers in neuroscience},
  volume={10},
  pages={508},
  year={2016},
  publisher={Frontiers Media SA}
}

@inproceedings{shrestha2018slayer,
  title={Slayer: Spike layer error reassignment in time},
  author={Shrestha, Sumit B and Orchard, Garrick},
  booktitle={Proceedings of the International
Conference on Neural Information Processing Systems (NeurIPS)},
  volume={31},
  year={2018}
}

@article{lee2020enabling,
  title={Enabling spike-based backpropagation for training deep neural network architectures},
  author={Lee, Chankyu and Sarwar, Syed Shakib and Panda, Priyadarshini and Srinivasan, Gopalakrishnan and Roy, Kaushik},
  journal={Frontiers in neuroscience},
  volume={14},
  pages={119},
  year={2020},
  publisher={Frontiers}
}

@article{neftci2019surrogate,
  title={Surrogate gradient learning in spiking neural networks: Bringing the power of gradient-based optimization to spiking neural networks},
  author={Neftci, Emre O and Mostafa, Hesham and Zenke, Friedemann},
  journal={IEEE Signal Processing Magazine},
  volume={36},
  number={6},
  pages={51--63},
  year={2019},
  publisher={IEEE}
}

@inproceedings{xiao2021training,
  title={Training feedback spiking neural networks by implicit differentiation on the equilibrium state},
  author={Xiao, Mingqing and Meng, Qingyan and Zhang, Zongpeng and Wang, Yisen and Lin, Zhouchen},
  journal={Proceedings of the International Conference on Neural Information Processing Systems (NeurIPS)},
  volume={34},
  pages={14516--14528},
  year={2021}
}

@inproceedings{zhang2022spiking,
  title={Spiking Transformers for Event-Based Single Object Tracking},
  author={Zhang, Jiqing and Dong, Bo and Zhang, Haiwei and Ding, Jianchuan and Heide, Felix and Yin, Baocai and Yang, Xin},
  booktitle={Proceedings of the IEEE/CVF Conference on Computer Vision and Pattern Recognition (CVPR)},
  pages={8801--8810},
  year={2022}
}

@inproceedings{zhang2022spike,
  title={Spike Transformer: Monocular Depth Estimation for Spiking Camera},
  author={Zhang, Jiyuan and Tang, Lulu and Yu, Zhaofei and Lu, Jiwen and Huang, Tiejun},
  booktitle={Proceedings of the European Conference on Computer Vision (ECCV)},
  year={2022}
}

@inproceedings{mueller2021spiking,
  title={Spiking Transformer Networks: A Rate Coded Approach for Processing Sequential Data},
  author={Mueller, Etienne and Studenyak, Viktor and Auge, Daniel and Knoll, Alois},
  booktitle={2021 7th International Conference on Systems and Informatics (ICSAI)},
  pages={1--5},
  year={2021},
  organization={IEEE}
}

\end{document}